\pdfoutput=1

\documentclass[11pt]{article}

\usepackage[]{emnlp2021}

\usepackage{times}
\usepackage{latexsym}

\usepackage[T1]{fontenc}

\usepackage[utf8]{inputenc}

\usepackage{microtype}

\title{DensE: An Enhanced Non-commutative Representation for Knowledge Graph Embedding with Adaptive Semantic Hierarchy}

\usepackage{algorithmicx,algorithm}
\usepackage{appendix}
\usepackage{hhline}
\usepackage{url}
\usepackage{multirow}
\usepackage{multicol, blindtext}
\usepackage{graphicx}
\usepackage{makecell}

\usepackage{soul}
\usepackage[utf8]{inputenc}
\graphicspath{ {./images/} }
\usepackage{amsmath}
\usepackage{booktabs}
\usepackage{amsfonts}       
\usepackage{nicefrac}       
\usepackage{lipsum} 
\usepackage[noend]{algpseudocode}
\usepackage{stmaryrd}

\usepackage{amssymb}
\usepackage{mathtools, cuted}

\usepackage{float} 
\usepackage{rotating}
\usepackage{adjustbox}  
\usepackage{blindtext}

\usepackage{subcaption}
\usepackage[usestackEOL]{stackengine}

\author{Haonan Lu\textsuperscript{\thanks{Corresponding Author. This work was done when the author was working in Huawei Technologies.}} \\
  OPPO \\Guangdong Mobile\\Telecommunications Co., Ltd. \\
  \texttt{luhaonan@oppo.com} \\\And
  Hailin Hu \\
  Huawei \\Technologies Co., Ltd. \\
  \texttt{huhailin2@huawei.com} \\\And
  Xiaodong Lin \\
  Department of\\Management Science\\and Information Systems, \\Rutgers University \\
  \texttt{lin@business.rutgers.edu} \\}

\begin{document}
\maketitle


\begin{abstract}
Capturing the composition patterns of relations is a vital task in knowledge graph completion. It also serves as a fundamental step towards multi-hop reasoning over learned knowledge. Previously, several rotation-based translational methods have been developed to model composite relations using the product of a series of complex-valued diagonal matrices. However, these methods tend to make several oversimplified assumptions on the composite relations, e.g., forcing them to be commutative, independent from entities and lacking semantic hierarchy. To systematically tackle these problems, we have developed a novel knowledge graph embedding method, named DensE, to provide an improved modeling scheme for the complex composition patterns of relations. In particular, our method decomposes each relation into an SO(3) group-based rotation operator and a scaling operator in the three dimensional (3-D) Euclidean space. This design principle leads to several advantages of our method: (1) For composite relations, the corresponding diagonal relation matrices can be non-commutative, reflecting a predominant scenario in real world applications; (2) Our model preserves the natural interaction between relational operations and entity embeddings; (3) The scaling operation provides the modeling power for the intrinsic semantic hierarchical structure of entities; (4) The enhanced expressiveness of DensE is achieved with high computational efficiency in terms of both parameter size and training time; and (5) Modeling entities in Euclidean space instead of quaternion space keeps the direct geometrical interpretations of relational patterns. Experimental results on multiple benchmark knowledge graphs show that DensE is comparable to the current state-of-the-art models for missing link prediction, especially on composite relations. In addition, the interpretations generated by DensE also reveal how relations with distinct patterns (i.e., symmetry/anti-symmetry, inversion and composition) are modeled, which suggests several important directions of future studies.
\end{abstract}

\section{Introduction}
Knowledge graphs (KGs) are a vital component of a wide range of downstream applications, such as machine reasoning, information retrieval and knowledge-guided natural language processing~\cite{ji2020survey, Zhang2019ERNIEEL, yang-etal-2019-enhancing-pre, Lin2019KagNetKG}. Especially, learning how to hop over a variety of concepts or instances stored in a knowledge graph represents a value path towards artificial general intelligence.

Knowledge graphs are defined as a collection of triplets. Each triplet, denoted by \((h, r, t)\), indicates a relation \(r\) pointing from the head entity $h$ to tail entity \(t\). Currently, numerous research efforts have been devoted to developing knowledge graph embedding (KGE) methods. These methods aim to learn a set of low-dimensional representations of entities and relations~\cite{ji2020survey,nguyen2017novel}, which is usually coupled with a score function to enable the knowledge graph completion process, i.e., predicting missing links between entities, for real-world KGs~\cite{nickel2016holographic, Lacroix2018CanonicalTD, bordes2013translating, sun2019rotate, zhang2019quaternion}. Sometimes, neural networks can be inserted into the process~\cite{dettmers2018convolutional,schlichtkrull2018modeling,nathani2019learning}, though this requires additional computation costs.


In principle, the desired KGE method should be able to accommodate various relation patterns and to learn representations that are approximately able to reason over the given patterns (expressiveness property of a KGE model~\cite{sun2019rotate, DBLP:journals/corr/abs-1709-04808}). 
For example, in a relation pattern such as symmetry (e.g., friend), asymmetry (e.g., uncle), inversion relations (e.g., hypernym and hyponym) and compositional relations (e.g., my father's mother is my grandmother), these patterns should be hold in the vector space.  
While the former three patterns are readily covered by the current methods~\cite{trouillon2016complex,sun2019rotate}, it still lacks an effective modeling strategy for composite relations due to the complexity of composition patterns. In particular, we find three predominant challenges in this modeling problem. First, the composition of relations can be non-commutative (e.g., my father's mother is my grandmother, while my mother's father is my grandfather, the reasoning result can be different by changing the orders of relations in a path of knowledge graph), which is opposite to the assumption of most KGE methods~\cite{bordes2013translating, sun2019rotate}. Second, the expressiveness of KGE methods is often limited by the counterintuitive lack of interaction between entity and relation embeddings~\cite{yang2014embedding}. Last but not least, while the semantic hierarchy of entities in a knowledge graph is a ubiquitous property~\cite{zhang2020learning}  (e.g., a triplet in WordNet~\cite{miller1995wordnet} \((palm, hypernym, tree)\) indicates ``tree'' is at a higher level than ``palm'' in the hierarchy), most methods do not pay attention to this and therefore fail to capture the semantic features at different semantic hierarchical levels.

\begin{figure*}[!ht]
\centering     
\begin{minipage}{9.0cm}
\centering
\includegraphics[width=90mm]{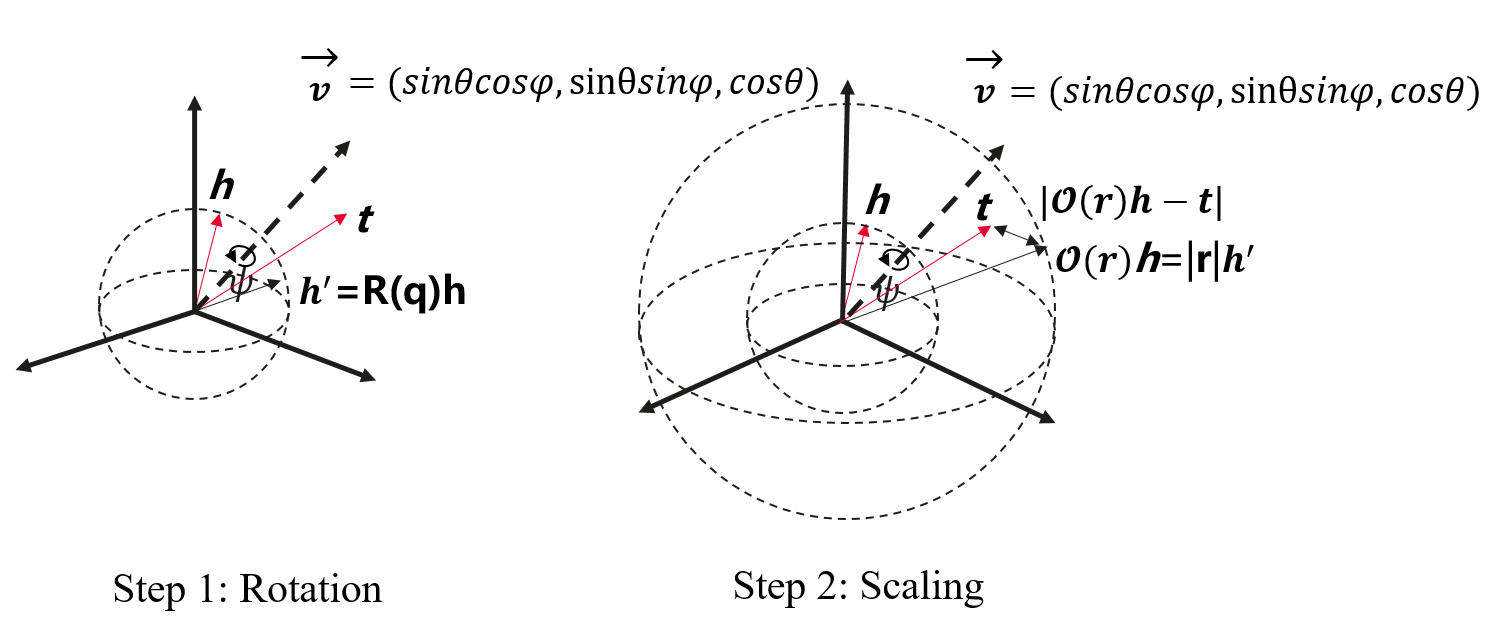}
\subcaption{}
\end{minipage}
\begin{minipage}{6.0cm}
\centering
\includegraphics[width=60mm]{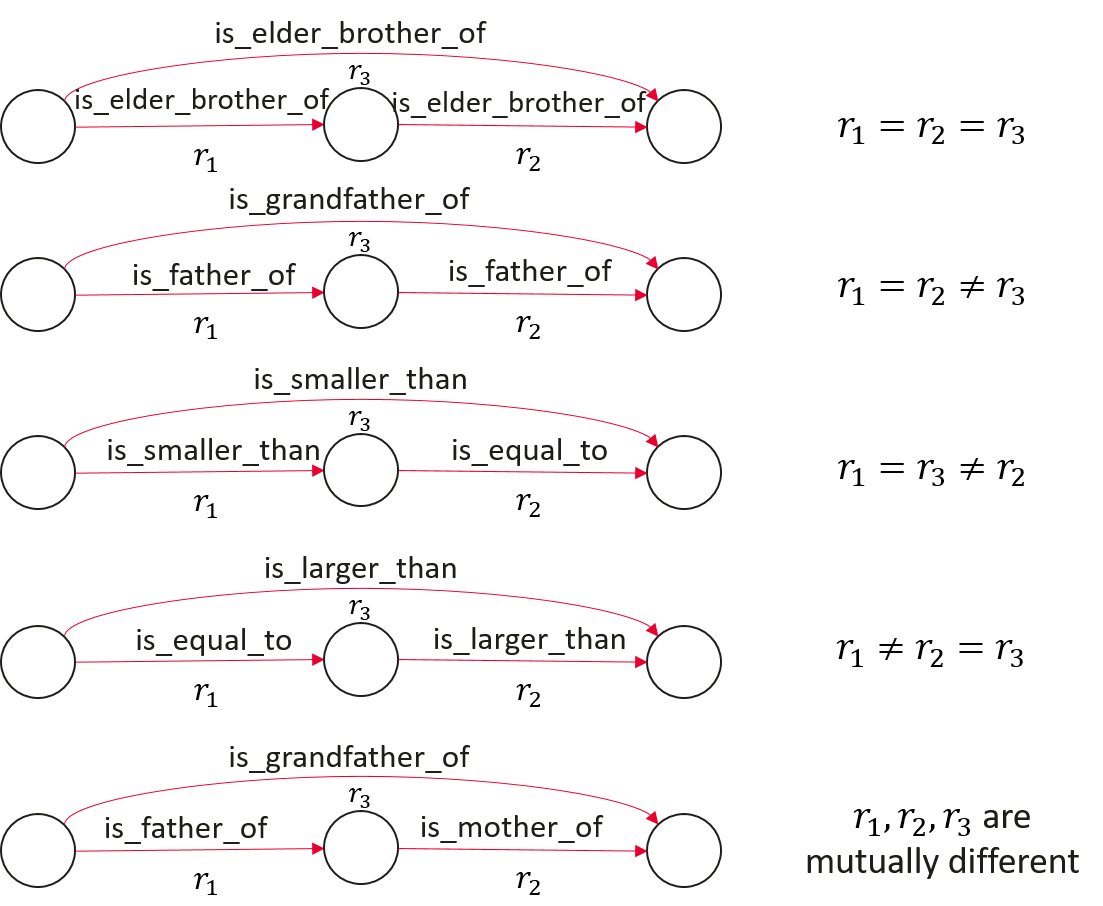}
\subcaption{}
\end{minipage}
\caption{(a) DensE decomposes a relation into a rotation operator and a scaling operator on the head entity \(\boldsymbol{h}\) in 3-D Euclidean space. (b) Examples of composition patterns.}
\label{comparison diffpool and graphstar}
\end{figure*}

In this work, to address these limitations, we develop a comprehensive solution to provide a highly expressive, efficient and interpretable modeling method for knowledge graph embedding. More specifically, we propose DensE (\textbf{D}istance-based \textbf{E}mbedding with \textbf{N}on-commutative Rotation and \textbf{S}caling in 3-D \textbf{E}uclidean Space), which decomposes the relation into an SO(3) group-based rotation operator and a scaling operator and in the 3-D Euclidean space. Intuitively, non-Abelian group (here we use SO(3) rotation group) is applied to introduce non-commutative nature to our model, and the scaling operation offers another important dimension to accommodate each triplet in the Euclidean space, which is barely explored in previous research. Our main contributions are summarized as the following: 

(1) By integrating infinite non-Abelian group-based relational rotation and scaling operations in the 3-D Euclidean space within a unified framework, we effectively accommodate various relation patterns including non-commutative compositions, semantic hierarchy, as well as interactions between entities and relations;

(2) Extensive experiments show that DensE achieves comparable to the current state-of-the-art models in link prediction with high computational efficiency, offering a useful tool for knowledge graph completion;

(3) We systematically consider three important scenarios of composition patterns that shall be considered by KGE methods (\textbf{Section~3}). Then, we show our method can provide an up-to-date most comprehensive while straightforward geometric interpretation for the modeling process of each relation type in the 3-D Euclidean space.

\section{Related Work}
In this section, we will discuss two different categories of KGE methods, especially how they evolve in terms of model expressiveness and interpretability.

\subsection{Transnational Distance Model}


Transnational distance models, represented by TransE~\cite{bordes2013translating} and RotatE~\cite{sun2019rotate} use Eucleadian distance as the score function. In particular, the embedding of relations and entity are fit so that the tail entity can be obtained by from the head entity using the operation defined by the relation. Given this intuition, these methods usually reflect some particular geometric interpretations. However, within the various relation patterns, their modeling capacity for composite relations (i.e., a relation path composed of a series of relations) tend to be insufficient because most methods assume a commutative pattern on the relation path and do not consider entity information in inferring composition patterns. 

Specifically, TransE models each relation as a pure translational transformation, so it assumes a fixed addition composition pattern between relations, i.e., \(r_3 = r_1 + r_2\), which is commutative and irrelevant to entity embeddings. RotatE made significant progress by modeling relations as rotational operator (rotation matrix) in 2-D Euclidean space. When modeling a relation path composed of multiple relations, RotatE uses Hadamard product to combine the rotation matrices of the relations on the path, i.e., \(r_3 = r_1 \circ r_2\). In this model, all relations in the composite relation have the same rotation axis. Thus, the compositions in RotatE are also mandatorily commutative. Also, interactions between relation and entity embeddings are precluded as the rotation axis is always perpendicular to entity embeddings.

Following the effort of RotatE, several methods have been proposed to enhance the expressiveness of rotation-based translation KGE model. For instance,~\cite{Yang2020AGF} proposes a group-theoretic analysis for KGE methods. Their method, named NagE, represents a preliminary attempt in applying non-Abelian group in modeling relational rotations. RotatE3D~\cite{gao2020rotate3d}, on the other hand, extends the rotation of RotatE into the 3-D Euclidean space. However, although these methods represent certain conceptual advances, their empirical results show limited performance advance over previous methods, probably challenged by the fitting power of pure rotation-based operations and lacking of ability to model semantic hierarchies in knowledge graphs. 

\subsection{Semantic Matching Model}


In constrast to translational distance model, methods in this category evaluate the matching of latent representation of relations and entities using bilinear model, e.g., RESCAL~\cite{nickel2011three}, DistMult~\cite{yang2014embedding}, and ComplEX~\cite{Trouillon2016ComplexEF}. 
Recently, as a generalization of DistMult and ComplEX, QuatE proposes a transformation on the entity representations by quaternion multiplication with the relation representation~\cite{zhang2019quaternion}, leading to a significant advance in expressiveness. For composite relations, this method does not assume any fixed composition pattern and preserve the non-commutative nature to some extent. However, QuatE requires normalization of relation to unit quaternion, indicating it is incapable of integrating scale information. In addition, as both entities and relations are embedded in quaternion hyperplanes, QuatE cannot provide a straightforward geometric interpretation in the space, which hinders the understanding of the learned embeddings. 



\subsection{Methodological Advance of DensE}
From the perspective of score function, DensE also belongs to translation distance model. In contrast to the previous works, our model leverages both rotation and scaling operations for relation modeling. The key idea is that we can transform any non-zero vector in the Euclidean space to another arbitrary vector through decoupled rotation and scaling transformations. In addition, our model provides a clear geometric picture to demonstrate the transformation of entity representation in various relation composition patterns.

Conceptually, some modules of our method is also related to other recent KGE methods. For instance, a recent work HAKE~\cite{zhang2020learning} has explored the combination of rotation and scaling operations, which are used to model  entities at same and different levels of hierarchy, respectively. In HAKE, rotation is defined following the protocol in RotatE, which leverages U(1) Abelian group and thus incapable of handling non-commutative relations. In addition, its rotation axis is vertical to the 2-D representation space of entities, again omitting the entity-relation interaction as in RotatE. On the other hand, while using the non-Abelian group in KGE model has been explored by~\cite{xu2019relation}, we argue this model has lower expressiveness than us in principle since it uses finite (non-)Abelian group (using non-Abelian group is optional) in 2-D space while we consider an infinite non-Abelian group in 3-D space. Also, its rotation operation is combined with reflection, which constitutes a special case of our scaling operation (i.e., scaling with a factor of -1). 

In contrast with QuatE that models both entities and relations in the quaternion space and does transformation using quaternion multiplication, DensE is based on 3-D Euclidean space rather than the space of quaternions. The continuous rotation transformation in the n-D Euclidean space (\(n > 2\)) is modeled by a special orthogonal group (SO(n) group). 
Compared with the vanilla U(1) abelian group based KGE models (e.g. RotatE/HAKE) that perform rotation transformation in the 2-D Euclidean space, continuous rotation transformation in the 3-D space modeled by SO(3) group is the minimum non-abelian extension with geometric interpretability. 
Composite relations (relation paths, details can be found in \textbf{Section 3}) are usually modeled by the product/summation of relation matrices. The violation of the commutative law of multiplication in the non-abelian case makes modeling the complex composition patterns of relations possible (non-commutative). 
The quaternion system is related to the SO(3) group and corresponds to the rotation transformation in 3-D Euclidean space, it provides a mathematical way to model continuous rotation transformation in 3D space. In our model, to guarantee geometric interpretability, entities are represented by 3D vectors, relations are modeled by quaternions that perform rotation and scaling transformation in 3D space. 
The rigidness of the quaternion system corresponds to the mathematical properties of SO(3) group theory since the rotation transformation in 3D space must satisfy several constraints such as: non-commutative (non-abelian nature), orthogonality, invertibility, etc. 
QuatE also studied that increasing spatial dimensions such as to Octonion does not increase performance compared to modeling relation and entities in the space of quaternion. The reason behind this is that the octonion system is more rigid than the quaternion system, the associative law of multiplication is also violated. However, this property is even harmful to modeling patterns of relations in a KG since there’s no relation pattern in a real-world KG that needs to be modeled by violating the associative law of multiplication.

Therefore, we argue that among these concurrent works, DensE is the most comprehensive solution with the geometric interpretability that accounts for all the three \textit{desiderata} for modeling composite relations, i.e., covering the non-commutative relations, preserving interaction between entity and relation, and capturing the entities' semantic hierarchy.

\section{Problem Formulation}

We denote a directed knowledge graph as \(G(\mathcal{E},\mathcal{R}, \mathcal{F})\), where \(\mathcal{E}\), \(\mathcal{R}\) and \(\mathcal{F}\) are sets of entities, relations and facts, respectively. A fact stored in a KG can be expressed as a triplet \((h, r, t) \in \mathcal{F}\), where \(h, t \in \mathcal{E}\) and \(r \in \mathcal{R}\). Herein, we focus on the knowledge completion task, which aims to predict missing links based on the observed facts. To fulfill this goal, a score function is used to measure the plausibility of proposed fact candidates, and the goal of model optimization is to give higher scores to true triplets \((h, r, t)\) than the false triplets \((h, r, \bar{t})\) or \((\bar{h}, r, t)\), where \(\bar{t}\) and \(\bar{h}\) are randomly sampled tail and head entities, respectively. Mathematically, the entity and relation embeddings are usually represented by tensors, and the score function can thus be written into the form of \(f_{\textbf{r}}(\textbf{h}, \textbf{t})\). 

In principle, KGE models should be designed to accommodate various relation patterns existing in real world KGs, such as symmetry, anti-symmetry, inversion and composition, which are formally defined as follows.



Let \(x, y, z\) be the entities in a given KG, and \(r(\cdot, \cdot)\) maps the relation between the two entities, we have:

{\textbf{Definition 1.} A relation \(r\) is \textbf{symmetric} if \(\forall x, y\),
\begin{equation}
r(x, y) \Rightarrow r(y, x).
\label{symmetric/anti-symmetry def}
\end{equation}
\noindent On the other hand, a relation is said to be \textbf{anti-symmetric} if \(\forall x, y\),
\begin{equation}
r(x, y) \Rightarrow \neg r(y, x).
\end{equation}}
\(Friend\) is a typical example of symmetric relation, which means if we know \(x\) is friend of \(y\), we can infer \(y\) is also friend of \(x\). \(Filiation\) is an example of anti-symmetric relation.

\textbf{Definition 2.} Relation \(r_1\) is \textbf{inverse} to relation \(r_2\) if \(\forall x, y\)
\begin{equation}
r_1(x, y) \Rightarrow r_2(y, x) 
\label{inverse_def_supp}
\end{equation}
For instance, \(has\_part\) and \(part\_of\) fit into the scope of inverse relations, which means if we know \(x\) is a part of \(y\), we can infer that \(y\) has part \(x\). Note that both \textbf{symmetric/antisymmetric} and \textbf{inverse} relation patterns can be inferred in one hop, so they are also called \textbf{atomic} relation.

In contrast to the above atomic relation patterns (inferable within one hop), the complex composition patterns pose a particular challenge to modeling, as discussed below.

\textbf{Definition 3.} Relation \(r_3\) is \textbf{composed} of relation \(r_1\) and relation \(r_2\) if \(\forall x, y, z\)
\begin{equation}
r_1(x, y)  \Lambda  r_2(y, z) \Rightarrow r_3(x, z) 
\label{composite_def}
\end{equation}
Here \(r_3\) is also referred to as a \textbf{composite relation} and possesses certain \textbf{composition pattern}. 
In particular, our model design takes the following properties into account:

\textbf{Property 1.} The two relations in the composition are not always \textbf{commutative}. For example, given \(r_1=is\_father\_of\), \(r_2=is\_mother\_of\), based on the \textbf{Definition 3}, we will get \(r_3=is\_grandmother\_of\). However, when we change the order, i.e., \(r_1=is\_mother\_of\), \(r_2=is\_father\_of\), we will get \(r_3=is\_grandfather\_of\). Recent KGE methods usually model composite relations (relation paths) by the product (e.g., QuatE, HAKE) /summation (e.g., TransE) of relation matrices. However, it is non-trivial to model composition relation patterns since the product/summation of diagonal real-valued/complex-valued matrices is usually commutative and hence invariant with the order of relations. For instance, ComplEx~\cite{trouillon2016complex} models relation path mentioned above \(r_1(x,y) \Lambda r_2(y, z)\) by using the product of two complex-valued diagonal matrices: \(R_{r_2}R_{r_1}\). However, the product of relation matrices in the diagonalized framework are commutative since that \(R_{r_2}R_{r_1} = R_{r_1}R_{r_2}\).

\textbf{Property 2.} The composition patterns are not always inferable by the relations alone. For example, given that \(y\) is \(x'\)s younger sister and \(z\) is \(y'\)s elder brother, we can not answer whether \(z\) is elder or younger than \(x\) from the given information. Actually, to answer this question, we need to know more about \(x/y/z\) from their own attributes and their other relationships.

\textbf{Property 3.} In a composition, the relations involved are not necessarily different. Given the two-hop example above, besides the situation that \(r_1\), \(r_2\) and \(r_3 \) are mutually different, there are also four different cases that satisfy the definition of composition, i.e., \(r_1 = r_2 = r_3 \), \(r_1 = r_2 \neq r_3 \), \(r_1 = r_3 \neq r_2 \) and \(r_1 \neq r_2 = r_3 \) (Figure~1(b)).

\section{Method}

In this section, we will first discuss the limitation of previous method such as RotatE, which is based on the 2-D Euclidean space. Then we will introduce each module of our method. In particular, we model a relation by a combination of an SO(3) group-based rotation (introducing the non-commutative nature) and a scaling operation (introducing the semantic hierarchy).

\subsection{Limitations of Modeling Relational Rotation in the 2-D Euclidean Space}

\begin{figure}[!ht]
\centering     
\centering
\includegraphics[width=4cm]{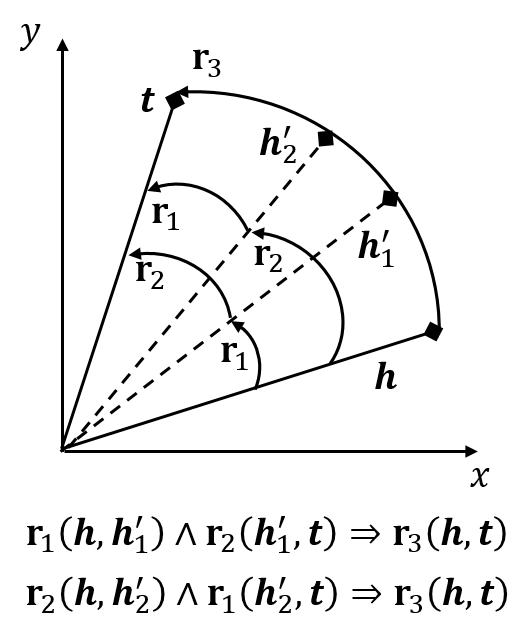}
\caption{RotatE models relations as a unit rotation operator in the 2-D Euclidean space.}
\label{rotation_in_2D}
\end{figure}

The motivation of RotatE is from Euler’s identity \(e^{i\theta} = \cos{\theta} + i\sin{\theta}\), which applies rotation in the 2D complex plane by using a unitary complex number. The RotatE model maps the entities and relations to the complex vector space and defines each relation as a rotation operator that transforms the source entity to the target entity. However, as shown in the Figure~\ref{rotation_in_2D}, composite relations are assumed to be commutative. Changing the order of relational rotation of \(r_1\) and \(r_2\) gives the same composition \(r_3\). Also, the unit rotation transformation makes it difficult to model the semantic hierarchy which is a ubiquitous property in knowledge graphs. The rotation axis (perpendicular to the paper) of 2-D rotation transformation is orthogonal to entity embeddings, which hinders the method to model interactions between relational operations and entity embeddings. 

Continuous rotation transformation in 3D space modeled by SO(3) group is the minimum non-abelian extension with geometric interpretability. By modeling relations and entities as rotation operators and vectors in 3D space, transformation in the 3-D Euclidean space can be either non-commutative or commutative. And since the rotation axis of 3-D transformation is not enforced to be perpendicular to the vectors, interactions between relations and entities can also be considered. A simple example for how rotation in the 3-D Euclidean space can model non-commutative relations is shown in \textbf{Supplementary Note 1}. We will formally introduce the mathematical method for modeling rotations in 3D space in \textbf{Section 4.2}. In addition, to model the semantic hierarchies of knowledge graphs, the rotation operation is then followed by a scaling transformation. The modulus parts of 3D vectors aims to model the entities in a KG at different levels of the semantic hierarchy. The detail of integrating the scaling transformation in our model is discussed in \textbf{Section 4.3}.


\subsection{Modeling Relational Rotation Using SO(3) Rotation Group}

One of the ways to model a rotation operation in the 3-D space is called axis-angle representation, which parameterizes a rotation by two quantities: 1) A unit vector \(\overrightarrow{\boldsymbol{v}}\) indicating the direction of the axis of rotation, i.e.,  \(\overrightarrow{\boldsymbol{v}} = (v_x, v_y, v_z) = (\sin{\theta}\cos{\phi}, \sin{\theta}\sin{\phi}, \cos{\theta})\), where \(\theta \in [0, \pi]\) and \(\phi \in [0, 2\pi)\); and  2) An angle \(\psi\) describing the magnitude of the rotation about the rotation axis, where \(\psi \in [0, 2\pi)\). Given an entity vector \(\overrightarrow{\boldsymbol{w}}\) in the 3-D space with the coordinate \((x, y, z)\), its rotation about axis \(\overrightarrow{\boldsymbol{v}}\) with an angle of \(\psi\) can be modeled using the SO(3) group theory (Figure~1(a), Step 1). More specifically, we can use a unit quaternion to encode the rotation using three degrees of freedom (i.e., \(\theta\), \(\phi\) and \(\psi\)). Actually, it can be viewed as a group structure on a 3-sphere (i.e., S3) which gives the group Spin(3). Note that this group structure is isomorphic to SU(2) group and also to the universal cover of SO(3) group. Formally, the unit quaternion \textbf{q} to model a rotation through an angle of \(\psi\) around the aforementioned axis \(\overrightarrow{\boldsymbol{v}}\) can be derived using an extension of Euler's formula:
\begin{equation}
\textbf{q} = e^{\frac{\psi}{2}(v_x\textbf{i} + v_y\textbf{j} + v_z\textbf{k})} = \cos{\frac{\psi}{2}} + \sin{\frac{\psi}{2}} * (v_x\textbf{i} + v_y\textbf{j} + v_z\textbf{k}),
\label{unit_quat_def}
\end{equation}
\noindent where \(\textbf{i}, \textbf{j}, \textbf{k}\) are imaginary units of the quaternion representation, which satisfies the condition \(\textbf{i}^2 = \textbf{j}^2 = \textbf{k}^2 = \textbf{ijk} = -1\).
Unlike real/complex numbers, the multiplication of quaternions (Hamilton product) is sensitive to the orders as we have: \(\textbf{ij}=\textbf{k}, \textbf{ji}=-\textbf{k}, \textbf{jk}=\textbf{i}, \textbf{kj}=-\textbf{i}, \textbf{ki}=\textbf{j}, \textbf{ik}=-\textbf{j}\).
For \(\textbf{Q}_1 = a_1 + b_1\textbf{i} + c_1\textbf{j} + d_1\textbf{k}\) and \(\textbf{Q}_2 = a_2 + b_2\textbf{i} + c_2\textbf{j} + d_2\textbf{k}\), their Hamilton product is:
\begin{equation}
\begin{aligned}
\textbf{Q}_1 \otimes \textbf{Q}_2 = a_1a_2 - b_1b_2 - c_1c_2 - d_1d_2 \\ + (a_1b_2 + b_1a_2 + c_1d_2 - d_1c_2)\textbf{i} \\
+ (a_1c_2 - b_1d_2 + c_1a_2 + d_1b_2)\textbf{j} \\
+ (a_1d_2 + b_1c_2 - c_1b_2 + d_1a_2)\textbf{k}
\label{quat_prod}
\end{aligned}
\end{equation}

A 3-D Euclidean vector \(\overrightarrow{\boldsymbol{w}}\) with the coordinate \((x, y, z)\) can be expressed as a pure quaternion (meaning the real part of quaternion is zero), i.e., \(\textbf{W} = x\textbf{i} + y\textbf{j} + z\textbf{k}\), giving the following theorem~\cite{jia2019quaternions}:
\textbf{Theorem 1} Given a 3-D Euclidean vector  \(\overrightarrow{\boldsymbol{w}}\) and its counterpart in the quaternion space \(\textbf{W}\), the desired rotation axis \(\overrightarrow{\boldsymbol{v}}\), the magnitude of the rotation \(\psi\), the destination coordinate of the vector after the rotation, i.e.,\(\textbf{W}' = x'\textbf{i} + y'\textbf{j} + z'\textbf{k}\), can be calculated by the Hamilton product of quaternions:
\begin{equation}
\begin{aligned}
\textbf{W}' = \textbf{q}\textbf{W}\textbf{q}^{-1}
\label{rotation_quad}
\end{aligned}
\end{equation}
\noindent where \(\textbf{q}^{-1}\) is the inverse of \(\textbf{q}\), i.e., \(\textbf{q}^{-1} = e^{-\frac{\psi}{2}(v_x\textbf{i} + v_y\textbf{j} + v_z\textbf{k})} = \cos{\frac{\psi}{2}} - \sin{\frac{\psi}{2}} * (v_x\textbf{i} + v_y\textbf{j} + v_z\textbf{k})\).

The form of Eq.\ref{rotation_quad} and a factor of \(\frac{1}{2}\) for the angle \(\psi\) in Eq.\ref{unit_quat_def} indicate that there is a \(2:1\) homomorphism from quaternions of unit norm to SO(3). Considering each 3-D Euclidean vector can also be expressed as a pure quaternion, we can now represent the rotation using a matrix \(\textbf{R(q)}\) by expanding Eq.\ref{rotation_quad} and letting \(C=\cos{\psi}\) and \(S=\sin{\psi}\):
\begin{scriptsize}
\begin{equation}
\begin{aligned}
& \overrightarrow{\boldsymbol{w}}' = \textbf{R(q)}\overrightarrow{\boldsymbol{w}} = \\ 
& \left[
 \begin{matrix}
   C+v_{x}^{2}(1-C) & v_{x}v_{y}(1-C)+v_{z}S & v_{x}v_{z}(1-C)-v_{y}S \\
   v_{x}v_{y}(1-C)-v_{z}S & C+v_{y}^{2}(1-C) & v_{y}v_{z}(1-C)+v_{x}S \\
   v_{x}v_{z}(1-C)+v_{y}S & v_{y}v_{z}(1-C)-v_{x}S & C+v_{z}^{2}(1-C)
  \end{matrix}
  \right] 
\left[
 \begin{matrix}
   x \\
   y \\
   z
  \end{matrix}
  \right] 
\label{matrix_cal}
\end{aligned}
\end{equation}
\end{scriptsize}

In our framework, two rotations can be combined into one equivalent rotation operation (this is also consistent with the closure property of group theory). In other words, we can define \(\textbf{q} = \textbf{q}_2\textbf{q}_1\), where \(\textbf{q}\) corresponds to the rotation \(\textbf{q}_1\) followed by the rotation \(\textbf{q}_2\). Therefore, a series of rotations can be \textbf{composed} together and then applied as a single rotation. Note that quaternion multiplication is not commutative unless \(\textbf{q}_1\) and \(\textbf{q}_2\) share the same rotation axes (i.e., \(\overrightarrow{\boldsymbol{v}}_1 = \overrightarrow{\boldsymbol{v}}_2\)), which can be seen from Eq.\ref{quat_prod}. This makes it possible to model both commutative and non-commutative relation patterns. 

\subsection{Integrating the Scaling Operation}

In a knowledge graph, different entities may have different level of semantic hierarchy given a particular relation. For example, in WN18RR, \textit{trade} is a \textit{hypernym} of \textit{transaction}, and \textit{man} is recorded to \textit{has\_part} to be \textit{arm}. In these cases, the head entity and tail entity show different abstraction levels or showing inclusion relationships. Intuitively, we argue that the difference of semantic hierarchy can be reflected by the scale of entity, as the entities possessing same level of abstraction tend to be achieved through rotation operations.

To define this intuition mathematically, we first obtain of norm of quaternions. Following Eq.\ref{unit_quat_def} and letting \textbf{q} to be the unit quaternion, an arbitrary quaternion with non-unit norm can be written as: \(\textbf{Q} = a + b\textbf{i} + c\textbf{j} + d\textbf{k} = |Q|\textbf{q}\), with the norm given by
\begin{equation}
\begin{aligned}
|Q| = \sqrt{a^2+b^2+c^2+d^2}
\end{aligned}
\end{equation}
\noindent where 
\begin{equation}
\begin{gathered}
a=|Q|\cos{\frac{\psi}{2}}, \\ b=|Q|\sin{\frac{\psi}{2}}\sin{\theta}\cos{\phi},\\ c=|Q|\sin{\frac{\psi}{2}}\sin{\theta}\sin{\phi}, \\
d=|Q|\sin{\frac{\psi}{2}}\cos{\theta}.
\end{gathered}
\end{equation}
By multiplying a scalar \(|Q|\) in the Eq.\ref{matrix_cal}, we can further introduce length as another degree of freedom to better match the ground-truth tail embedding vector (Figure~1(a), Step 2). Formally, we have:
\begin{equation}
\begin{aligned}
\overrightarrow{\boldsymbol{w}}' = |Q|\textbf{R(q)}\overrightarrow{\boldsymbol{w}} = \mathcal{O}(\textbf{Q})\overrightarrow{\boldsymbol{w}}, \text{ where } \textbf{Q} \in \mathbb{H}, \overrightarrow{\boldsymbol{w}}, \overrightarrow{\boldsymbol{w}}' \in \mathbb{R}^{3},
\label{final_math_form}
\end{aligned}
\end{equation}
\noindent where \(\mathcal{O}(\textbf{Q}) = |Q|\textbf{R(q)}\) is the combined operator of rotation and scaling transformations, \(\mathbb{H}\) denotes the quaternion algebra, and \(\mathbb{R}^{3}\) represents the 3-D Euclidean algebra. Here we call \(|Q|\) the \(scaling \text{ } factor\). Therefore, we now have a uniform framework with  interpretable geometric meaning, i.e., \((|Q|, \theta, \phi, \psi)\) to describe the transformation corresponding to a specific relation type.
We can also define the reverse operation \(\mathcal{O}(\textbf{Q}^{-1}) = |Q|^{-1}\textbf{R}(\textbf{q}^{-1})\), which describes the reverse process: rotate a vector about the axis \(\overrightarrow{\boldsymbol{v}}\) with angle \(-\psi\) (from another direction), and then scale the vector with a factor of \(|Q|^{-1}\). Combining the Eq.\ref{matrix_cal} and Eq.\ref{final_math_form}, we can always find a operator \(\mathcal{O}(\textbf{Q}_3)\) = \(\mathcal{O}(\textbf{Q}_2)\mathcal{O}(\textbf{Q}_1)\), which corresponds to the application of \(\mathcal{O}(\textbf{Q}_1)\) followed by the application of \(\mathcal{O}(\textbf{Q}_2)\), where we have \(|Q_3| = |Q_1|*|Q_2|\) and \(\textbf{R}(\textbf{q}_3) = \textbf{R}(\textbf{q}_2)\textbf{R}(\textbf{q}_1)\).

\subsection{Score Function and Optimization}
A score function aims to correctly measure the plausibility of a triple of interest. Formally, as a distance-based model, our scoring function is defined as 
\begin{equation}
f_{\textbf{r}}(\textbf{h}, \textbf{t}) = -\frac{1}{2}(|\mathcal{O}(\textbf{r})\textbf{h} - \textbf{t}| + |\mathcal{O}(\textbf{r}^{-1})\textbf{t} - \textbf{h}|). 
\end{equation}
\noindent Here, \(|\cdot|\) denotes the Euclidean distance and \(\mathcal{O}({\cdot})\) stands for the transformation conducted on each element of the entity embeddings. That is to say, for the \(i\)-th embedding unit of \(\textbf{h}\), the optimization target is to minimize the Euclidean distance between \(\boldsymbol{t}_i\) and \(\mathcal{O}(\textbf{r}_{i})\boldsymbol{h}_i\), as well as the Euclidean distance between \(\boldsymbol{h}_i\) and \(\mathcal{O}(\textbf{r}_i^{-1})\boldsymbol{t}_i\),
\noindent where \(\textbf{r}_i \in \mathbb{H}, \boldsymbol{h}_i, \boldsymbol{t}_i \in \mathbb{R}^{3}.\) \(\mathbb{H}\) and \(\mathbb{R}^{3}\) stand for the quaterion and 3-D Euclidean algebra, respectively. The arrow of \(\boldsymbol{h}_{i}\) and \(\boldsymbol{t}_{j}\) are omitted for clarity.
To properly train the model parameters, here we use a loss function similar to the self-adversarial negative
sampling loss proposed in~\cite{sun2019rotate}:
\begin{equation}
\begin{split}
L = -\log{\sigma(\gamma + f_{\textbf{r}}(\textbf{h}, \textbf{t}))} \\ - \sum_{j=1}^{n}{p(\bar{h}_j, r, \bar{t}_{j})\log{\sigma(-(\gamma + f_{\textbf{r}}(\bar{\textbf{h}}^{(j)},\bar{\textbf{t}}^{(j)})))}},
\label{loss}
\end{split}
\end{equation}
where \(\gamma\) is a fixed margin, \(n\) is the number of negative sampling size,  \((\bar{h}_j, r, \bar{t}_{j})\) is the \(j\)-th negative triplet of the fact \((h, r, t)\), and \(\sigma\) is the sigmoid function. 
\(\bar{\textbf{h}}^{(j)}\) and \(\bar{\textbf{t}}^{(j)}\)are the embeddings corresponding to the negative triplet \((\bar{h}_j, r, \bar{t}_{j})\).
\(p(\bar{h}_j, r, \bar{t}_{j})\) is the weight of the negative sample, which gives the higher scored negative samples with larger weight during training. The details about self-adversarial negative sampling technique can be found in~\cite{sun2019rotate}.

\section{Experimental Settings}

\begin{table*}[]
\centering
\caption{Statistics of datasets used in this study.}

\begin{tabular}{ l c c c c c  }
 \toprule
 Dataset  & \# Entities & \# Relations & \#{Training}  &  \#{Validation}   & \#{Test}  \\  
 \midrule
 WN18RR & 40943   & 11  & 86835  & 3034  & 3134   \\
 FB15k-237 & 14541   & 237  & 272115   & 17535   & 20466\\ YAGO3-10    & 123182    & 37  & 1079040   & 5000   & 5000 \\ 
\bottomrule
\end{tabular} 
\label{Stat}
\end{table*}

\begin{table*}[!ht]
\centering
\caption{Performance comparison on benchmark datasets. Best results are labeled in bold and the second best are underlined. The reporting scheme generally follows that in~\cite{ruffinelli2019you}. \textit{First} indicates the originally reported performance of each method. \textit{Enhanced} records the improved performance with tuned training techniques and hyperparameters by~\cite{ruffinelli2019you}. \textit{Recent} shows the best results of more selected recent models. \textit{Adv+Recip} reports the model performance using the same training scheme of DensE, i.e., using self-adversarial negative sampling and reciprocal learning. \textit{Ours} reports the performance of DensE as well as its ablation counterparts.}

 \begin{tabular}{cccccccc}
  \toprule
    \multicolumn{2}{c}{} & \multicolumn{2}{c}{\textbf{WN18RR}} & \multicolumn{2}{c}{\textbf{FB15K-237}} & \multicolumn{2}{c}{\textbf{YAGO3-10}} \\
   	&	Model & MRR	&	H@10 &	MRR	&	H@10	&	MRR	&	H@10	\\
  \midrule

    \multirow{4}*{\textit{First}}	&	RESCAL~\cite{wang2018evaluating}	&	0.420	&	0.447	&	0.270	&	0.427	&	-	&	-	\\
    ~	&	TransE~\cite{nguyen2017novel}	&	0.226	&	0.501	&	0.294	&	0.465	&	-	&	-	\\
    ~	&	DistMult~\cite{dettmers2018convolutional}	&	0.430	&	0.490	&	0.241	&	0.419	&	0.340	&	0.540	\\
    ~	&	ComplEx~\cite{dettmers2018convolutional}	&	0.440	&	0.510	&	0.247	&	0.428	&	0.360	&	0.550	\\
    \midrule															
    \multirow{4}*{\textit{Enhanced}} 	&	RESCAL~\cite{ruffinelli2019you}	&	0.467	&	0.517	&	\underline{0.357}	&	0.541	&	-	&	-	\\
    ~	&	TransE~\cite{ruffinelli2019you}	&	0.228	&	0.520	&	0.313	&	0.497	&	-	&	-	\\
    ~	&	DistMult~\cite{ruffinelli2019you}	&	0.452	&	0.531	&	0.343	&	0.531	&	-	&	-	\\
    ~	&	ComplEx~\cite{ruffinelli2019you}	&	0.475	&	0.547	&	0.348	&	0.536	&	-	&	-	\\
    \midrule															
    \multirow{8}*{\textit{Recent}} 	&	RotatE~\cite{sun2019rotate}	&	0.476	&	0.571	&	0.338	&	0.533	&	0.495	&	0.670	\\
    ~	&	NagE~\cite{Yang2020AGF}	&	0.477	&	0.574	&	0.340	&	0.530	&	-	&	-	\\
    ~	&	QuatE~\cite{jia2019quaternions}	&	0.481	&	0.564	&	0.311	&	0.495	&	-	&	-	\\
    ~	&	D4-STE~\cite{xu2019relation} 	&	0.480	&	0.536	&	0.320	&	0.502	&	0.472	&	0.643	\\
    ~	&	TuckER~\cite{balavzevic2019tucker}	&	0.470	&	0.526	&	\textbf{0.358}	&	\textbf{0.544}	&	-	&	-	\\
    ~   &   Rotate3D~\cite{gao2020rotate3d} & 0.489 & 0.579 & 0.347 & \underline{0.543} & - & -\\ 
    ~   &   \(\text{HAKE}^{1}\)~\cite{zhang2020learning} & \textbf{0.497} & \underline{0.584} & 0.336 & 0.533 & 0.522 & \underline{0.693}\\
    ~   &   \(\text{HAKE}^{2}\)~\cite{zhang2020learning} & \textbf{0.497} & 0.582 & 0.346 & 0.542 & \textbf{0.545} & \textbf{0.694}\\
    \midrule															
    \multirow{4}*{\textit{Adv+Recip}} 	&	TransE	&	0.230	&	0.535	&	0.330	&	0.525	&	0.460	&	0.661	\\
    ~	&	DistMult	&	0.444	&	0.533	&	0.316	&	0.497	&	0.427	&	0.627	\\
    ~	&	ComplEx	&	0.475	&	0.559	&	0.334	&	0.525	&	0.510	&	0.681	\\
    ~	&	RotatE	&	0.478	&	0.567	&	0.337	&	0.531	&	0.497	&	0.676	\\
    \midrule															
    \multirow{4}*{\textit{Ours}}	&	DensE	&	\underline{0.492}	&	\textbf{0.586}	&	0.351	&	\textbf{0.544}	&	\underline{0.541}	&	0.678	\\
    ~ & -Scaling		&	0.475	&	0.562	&	0.335	&	0.527	&	0.486	&	0.642 \\
    ~ & -Reciprocal	&	0.487	&	0.572	&	0.343	&	0.527   &	0.530	&	0.667		\\
    ~ & -Adv	&	0.486	&	0.572	&	0.306	&	0.481	&	0.452	&	0.642	\\

    
 \bottomrule
\end{tabular} 
\label{main results table}
\end{table*}

\paragraph{Datasets and evaluation metrics} The experiments are conducted mainly on three commonly used benchmark datasets, including WN18RR, FB15k-237 and YAGO3-10. 
As pointed out by ~\cite{Toutanova2015ObservedVL, dettmers2018convolutional}, WN18 and FB15k suffer from the test set leakage problem. One can predict missing links and attain the state-of-the-art results even using a simple rule-based model. To avoid this issue, two much more challenging datasets (WN18RR~\cite{dettmers2018convolutional} and FB15k-237~\cite{toutanova2015observed}) were released.
WN18RR comes from WordNet~\cite{miller1995wordnet}, compared with the previous version WN18, it removes inverse relations to provide a more realistic KGE method benchmark. Similarly, the FB15k-237 dataset is also extracted from the original Freebase dataset FB15K~\cite{bordes2013translating} by removing inverse relations. In addition, we also use the YAGO3-10~\cite{mahdisoltani2013yago3} dataset, which consists of a large collection of triplets from multilingual Wikipedia. These three datasets aim to assess the model performance on composition patterns. The main relation patterns of them are symmetry/anti-symmetry and composition. 
The basic statistics of the datasets are provided in Table~\ref{Stat}. Here, we report mean reciprocal rank (MRR) and Hits at 10 (H@10) for evaluation (the higher, the better), which is consistent with~\cite{ruffinelli2019you}. Other performance metrics are provided in the Supplementary Material.

\paragraph{Baselines} We mainly compare DensE with top-performing baseline models for KG link prediction, including both translational model and semantic matching model. As the early implementation of the baseline models may lack thorough configuration tuning or advanced learning techniques, direct compassion with these performances (denoted as \textit{First}) may be biased to later methods. Therefore, for early models such as RESCALL~\cite{nickel2011three}, TransE~\cite{bordes2013translating}, DistMult~\cite{yang2014embedding} and ComplEx~\cite{trouillon2016complex}, we also provide two improved versions, including \textit{Enhanced}, which was obtained through a sophisticated hyperparameter tuning procedure by~\cite{ruffinelli2019you} and~\textit{Adv+Recip}, which is obtained by us using the same self-adversarial negative sampling and reciprocal learning as for DensE. These two improved versions are provided to prompt the fairness of the comparison.

In addition to these methods, we also compare our model with more recently proposed KGE models (denoted as \textit{Recent}), such as RotatE~\cite{sun2019rotate}, QuatE~\cite{jia2019quaternions}, D4-STE~\cite{xu2019relation}, TuckER~\cite{balavzevic2019tucker}, Rotate3D~\cite{gao2020rotate3d} and HAKE~\cite{zhang2020learning}. All these recent methods have included some advanced training techniques similar to~\cite{ruffinelli2019you}.  


\begin{figure*}[!ht]
\centering     
\begin{minipage}{7cm}
\centering
\includegraphics[width=70mm]{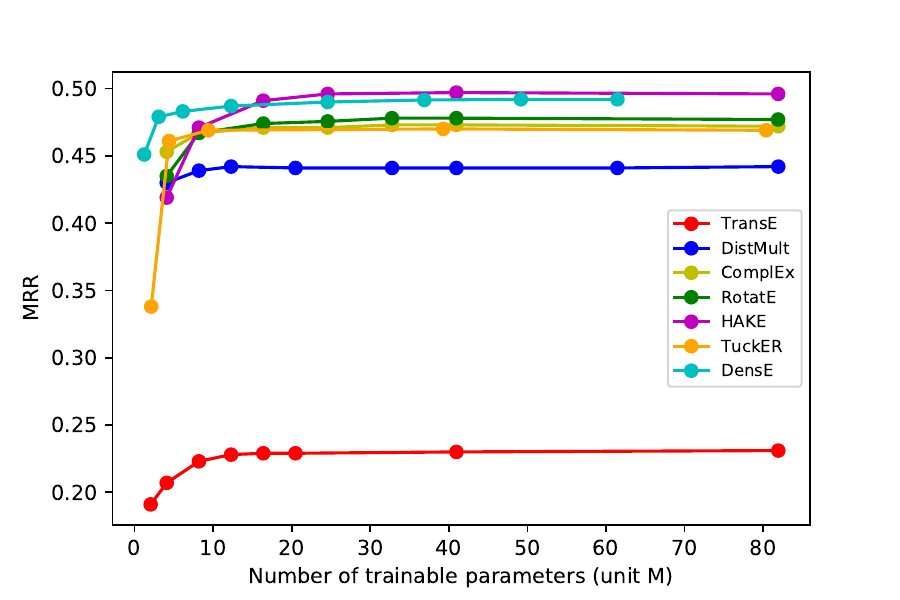}
\subcaption{}
\end{minipage}
\begin{minipage}{7cm}
\centering
\includegraphics[width=70mm]{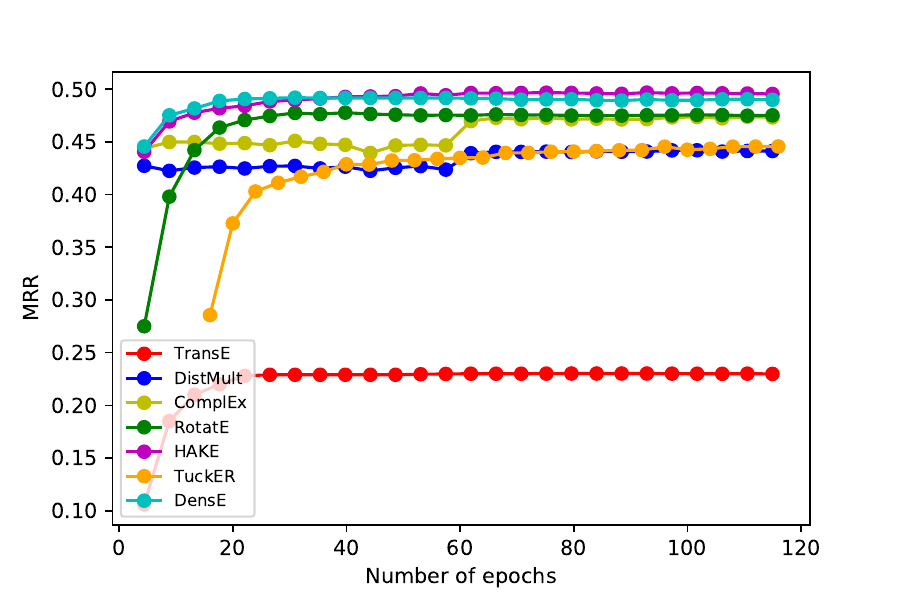}
\subcaption{}
\end{minipage}

\begin{minipage}{7cm}
\centering
\includegraphics[width=70mm]{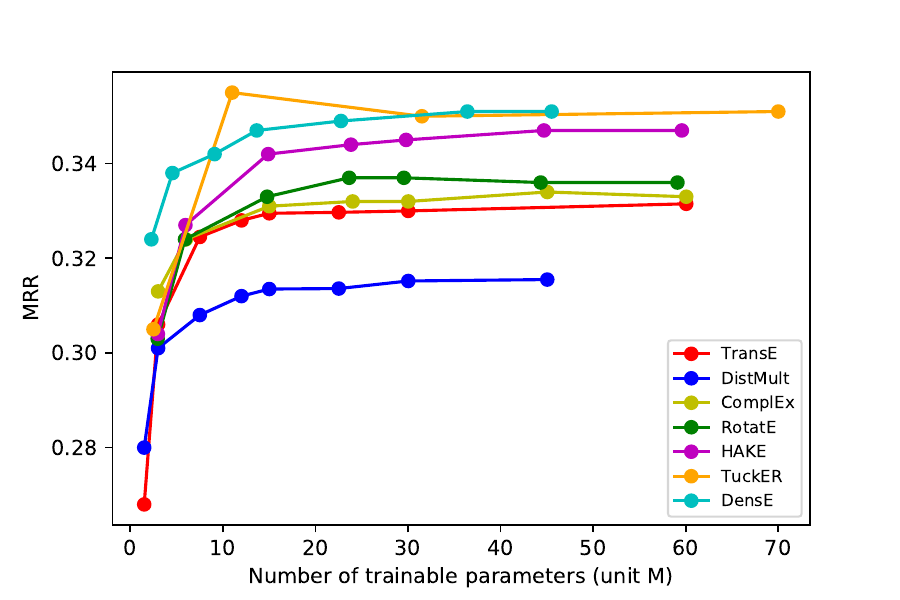}
\subcaption{}
\end{minipage}
\begin{minipage}{7cm}
\centering
\includegraphics[width=70mm]{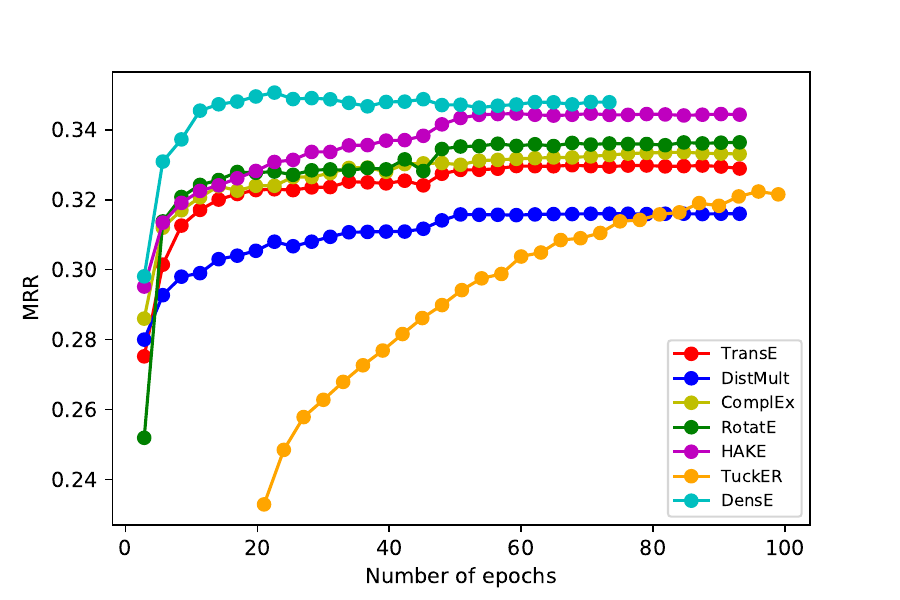}
\subcaption{}
\end{minipage}

\caption{The effect of trainable parameter size and training epoch number on model performances. (a) and (b) shows the results for WN18RR dataset. The corresponding results for FB15K-237 is shown in (c) and (d). All the results are achieved using the same setting ({\textit{Adv+Recip}}) as described above.}
\label{model_complexity}
\end{figure*}

\begin{figure*}[!ht]
\centering

\begin{minipage}{3.2cm}
\centering
\includegraphics[width=3.2cm]{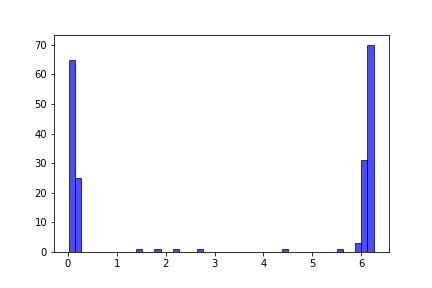}
\subcaption{\scriptsize{\(\psi_{has\_part} + \psi_{part\_of}\)}}
\end{minipage}
\begin{minipage}{3.2cm}
\centering
\includegraphics[width=3.2cm]{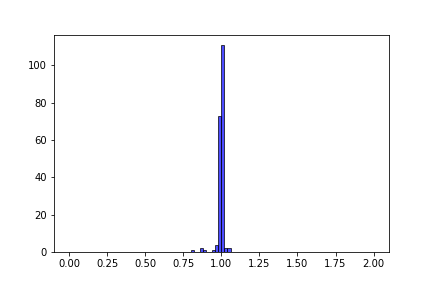}
\subcaption{\scriptsize{\(|Q_{has\_part}| * |Q_{part\_of}|\)}}
\end{minipage}
\begin{minipage}{3.2cm}
\centering
\includegraphics[width=3.2cm]{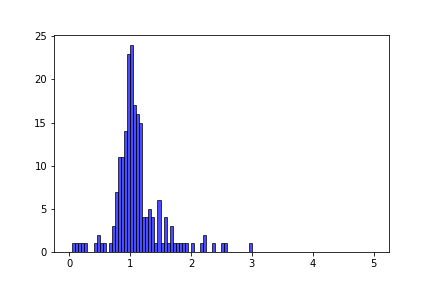}
\subcaption{\scriptsize{\(|Q_{has\_part}|\)}}
\end{minipage}%
\begin{minipage}{3.2cm}
\centering
\includegraphics[width=3.2cm]{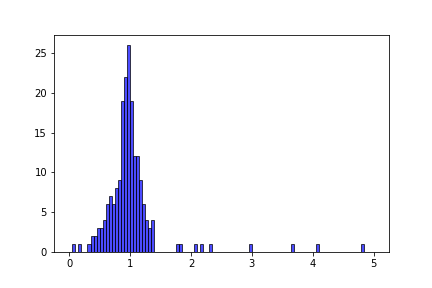}
\subcaption{\scriptsize{\(|Q_{part\_of}|\)}}
\end{minipage}

\begin{minipage}{4cm}
\centering
\includegraphics[width=3.2cm]{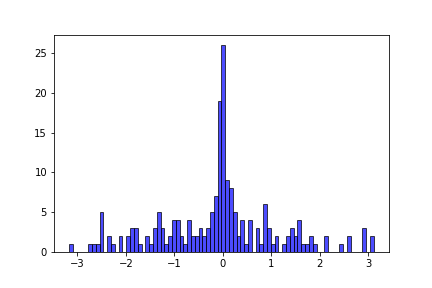}
\subcaption{\scriptsize{\(\psi(\mathcal{O}(\textbf{r2})\mathcal{O}(\textbf{r1}))-\psi(\mathcal{O}(\textbf{r1}))\)}}
\end{minipage}%
\hspace{0.5cm}
\begin{minipage}{4cm}
\centering
\includegraphics[width=3.2cm]{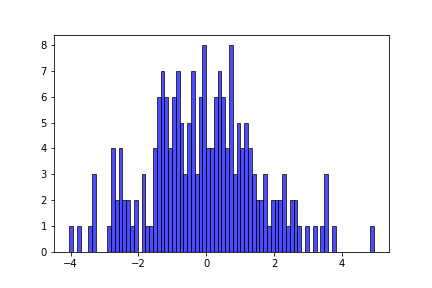}
\subcaption{\scriptsize{\(\psi(\mathcal{O}(\textbf{r2})\mathcal{O}(\textbf{r1}))-\psi(\mathcal{O}(\textbf{r3}))\)}}
\end{minipage}
\hspace{0.5cm}
\begin{minipage}{4cm}
\centering
\includegraphics[width=3.2cm]{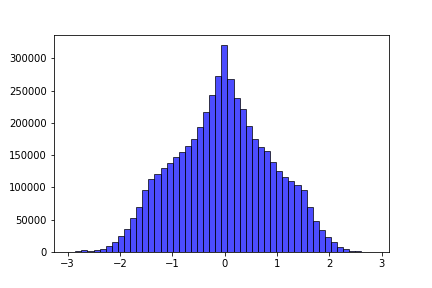}
\subcaption{\scriptsize{\(\theta(\mathcal{O}(\textbf{r1})) - \theta(\textbf{h})\)}}
\end{minipage}

\hspace{0.5cm}
\caption{Geometric interpretation provided by DensE. Each histogram shows a distribution of each dimension of the learned embeddings. Angular parameters are in radian units. (a)-(d) A case study for inversion patterns (Other degrees of freedom can be found in the \textbf{Supplementary Note 5}). (e)-(f) A case study for composition patterns, reflecting the scenario of \(r_1(h, h') \Lambda r_2(h', t) \Rightarrow r_1(h, t) \) and \(r_1(h, h') \Lambda r_2(h', t) \Rightarrow r_3(h, t) \), respectively. \(\psi(\cdot)\) denotes the rotation angle about the rotation axis of a relational operator. (g) Collinearity of entity and relation embedding. \(\theta(\mathcal{O}(\textbf{r1}))\) is the \(\theta\) component of relation \(r_1\). \(\theta(\textbf{h})\) is the \(\theta\) value of the head entity \(h\)'s embedding in the spherical coordinate system of the 3-D Euclidean space. All the head entities satisfying \(r_1(h, h') \Lambda r_2(h', t) \Rightarrow r_1(h, t) \) are included for the analysis.  The entities and relations involved can be found in the \textbf{Supplementary Note 5}.}
\label{Fig:Geo}
\end{figure*}

\paragraph{Implementation details} We use the Adam optimizer and tune the hyperparameters on the validation dataset. During training, we adopt a similar reciprocal learning approach as used in ~\cite{Lacroix2018CanonicalTD, zhang2019quaternion}. Early stopping is applied based on the performance on the validation dataset every 1,000 steps. The ranges for hyperparameter grid search and the best hyperparameter settings are listed in \textbf{Supplementary Note 2}. All the parameters are randomly initialized from the interval \\
\([-\frac{1}{\sqrt{2k}}, \frac{1}{\sqrt{2k}}]\), where \(k\) is the embedding size. 

\section{Results and Analysis}
\subsection{Prediction Performance} We report the link prediction results on the three benchmark datasets in Table~\ref{main results table}. 
On WN18RR, we show that DensE performs on par with HAKE and outperforms most other models on both the metrics, even after the baseline models are improved by hyper-parameter tuning or using advanced learning techniques. 
On FB15k-237, we show that the performance of DensE is superior to most of the baseline models, including RotatE and QuatE. While on this dataset we notice a particular good performance of a method called TuckER~\cite{balavzevic2019tucker}, this method also shows a significantly inferior performance on WN18RR, suggesting a potential drawback in generalizability. 
On YAGO3-10, DensE also shows a significant margin over RotatE and D4-STE (a KGE method based on 2D finite (non-)Abelian group), and also ComplEx when using the comprehensive metric MRR, which further demonstrates the superiority of DensE on various types of datasets. We also provide additional performance metrics, i.e., MR, MRR, Hits at 1 (H@1), Hits at 3 (H@3), and Hits at 10 (H@10) in Supplementary Note 3 (Supplementary Tables 2, 3, and 4).

Then we carefully compare the performance of DensE to two recent extension of RotatE models, namely RotatE3D~\cite{gao2020rotate3d} and HAKE~\cite{zhang2020learning}. We find that DensE performs better than Rotate3D in most cases, validating the contribution of the scaling operation. 
On the other hand, HAKE and DensE generally perform comparably. After dissecting into the training details, we find that different from DensE and most other translation distance model, HAKE pays more attention to model the hierarchical nature of knowledge graphs. \textbf{Firstly}, unlike the score function of DensE, which directly optimize the Euclidean distance between two vectors in the 3D space, HAKE decomposes the score function into two part: 1. The modulus-distance part corresponds to the hierarchy of the knowledge graph; 2. The phase-distance part corresponds to the rotation operation in the 2D space. HAKE leverages a task-specifically calibrated loss term by tuning the relative contributions of two terms in its score function manually to make it get better performance in the dataset with a clear hierarchical structure. Therefore, for a dataset like WN18RR with a majority of types of relations that link two entities at different levels of the hierarchy, HAKE can get better performance than DensE. The FB15k-237 dataset has more complex relation types (237 types of relation) and fewer entities (higher average degree of vertices) than WN18RR and YAGO3-10. The advantage of tuning relative contributions of two terms in the score function manually does not exist anymore, while the information of hierarchy can be learning automatically in our DensE with adaptive semantic hierarchy. That’s why we outperform HAKE in FB15k-237 dataset. \textbf{Secondly}, as mentioned in the HAKE, it has two versions of the score function to model the modulus-distance part. We report the results of two versions of HAKE in Table~\ref{main results table}, labeled by \(\text{HAKE}^{1}\) and \(\text{HAKE}^{2}\), respectively. Version 1 has a clear and simple mathematical form that models rotation in the 2D space. Compared with version 1, a bias and re-scaling operation on relational embedding are introduced into the model in version 2 (An additional freedom to tune embeddings of relation, thus the element of the embedding of relation is a 3D vector). For the YAGO3-10 dataset which is more complicated than WN18RR and also has a clear semantic hierarchy property, DensE outperforms \(\text{HAKE}^{1}\) and get comparable result with \(\text{HAKE}^{2}\). The above two techniques proposed by HAKE do improve the ability to model the semantic hierarchy and complement its lack of expressiveness in rotation operation. We believe these techniques will also be important tricks that can be used to improve performance in future studies (just like the self-adversarial negative sampling technique).

To confirm the source of performance gains, we conduct a further analysis that compares the MRR performance of DensE to RotatE on each relation type of WN18RR (Table~\ref{Breakdown}). 
Besides the taxonomy mentioned in \textbf{Section 3}, relations in the WN18RR dataset can be also divided into two categories: (a) relations that link two entities in the same semantic hierarchy (e.g., ``similar\_to''); (b) relations that link two entities at different levels of the hierarchy (e.g., ``has\_part''). One can see that most of the relations that fall into category (b) are also overlap with composite relation patterns. 
Intriguingly, we notice a large performance increase in composite relations, as exemplified by \(hypernym\), the most abundant composite relation in test data. We show that DensE improves MRR on this relation by as much as 3.3\%. 
These results indicates particular advantages of DensE in modeling composition relation patterns and semantic hierarchies of knowledge graphs. 


\begin{table*}[!h]
\centering
\caption{MRR comparison on each relation type of WN18-RR. Performance increases are in parentheses.}
 \begin{tabular}{c|cccc}
  \toprule
Relation type	&	Relation Name	&	\% in test data	&	RotatE	&	DensE	\\
  \midrule
\multirow{4}*{Atomic}	&	derivationally\_related\_form	&	34\%	&	0.947	&	0.955 (+0.008)	\\
~	&	also\_see	&	1.8\%	&	0.585	&	0.647 (+0.062)	\\
~	&	verb\_group	&	1.3\%	&	0.943	&	0.955 (+0.012)	\\
~	&	similar\_to	&	0.2\%	&	1	&	1 (+0)	\\
  \midrule
\multirow{7}*{Composite}	&	hypernym	&	39.5\%	&	0.148	&	0.181 (+0.033)	\\
~	&	instance\_hypernym	&	4\%	&	0.318	&	0.349 (+0.031)	\\
~	&	member\_meronym	&	8.1\%	&	0.232	&	0.249 (+0.017)	\\
~	&	synset\_domain\_topic\_of	&	3.8\%	&	0.341	&	0.412 (+0.071)	\\
~	&	has\_part	&	5.5\%	&	0.184	&	0.205 (+0.021)	\\
~	&	member\_of\_domain\_usage	&	0.8\%	&	0.318	&	0.326 (+0.008)	\\
~	&	member\_of\_domain\_region	&	1\%	&	0.2	&	0.407 (+0.207)	\\
  \bottomrule
\end{tabular} 
\label{Breakdown}
\end{table*}

\subsection{Ablation Study} To examine the effectiveness of each module in our model, we perform a series of ablation experiments (Table~\ref{main results table}). On WN18RR dataset, the most significant performance decrease occurs when we cancel the scaling operation, i.e., only model the relation as rotations. This confirms the contribution from scaling to the whole model. On FB15K-237 and YAGO3-10 datasets, we also observe a large drop in performance when removing the scaling operation. Also, self-adversarial negative sampling (adv) shows significant contribution, indicating the necessity to incorporate proper training techniques. In \textbf{Supplementary Note 4}, we also compared DensE and RotatE models without self-adversarial negative sampling and confirmed the superiority of DensE in this setting. Note that HAKE also uses the self adversarial technique in training. However, the ablation results are not provided.


\begin{table*}[h!]
\centering
\caption{The comparison of training time of each method on WN18RR. A lower total time results in a higher efficiency. To ensure a fair comparison, here we unify the hyperparateters of each method so that all the models have a similar parameter size around 36M.}

\begin{tabular}{ l c c c c }
 \toprule
 Model  & Training time per epoch (s) & \# of epochs & Total training time (s) & MRR  \\  
 \midrule
 DensE & 92 & 21 & 1932 & 0.492 \\
 RotatE & 83 & 40 & 3320 & 0.478 \\
 ComplEx & 75 & 62 & 4650 & 0.475 \\
 DisMult & 52 & 68 & 3536 & 0.444 \\
\bottomrule
\end{tabular} 
\label{time}
\end{table*}

\subsection{Computational Complexity} We show that compared with high-performance models such as RotatE and HAKE, DensE is generally more computationally efficient in terms of parameter number and training epochs. 
For TuckRE, although it can get better performance on FB15k237 with a relatively small model size, it needs much larger training epochs than other models on both WN18RR and FB15k237 datasets. We find that TuckRE needs roughly 500 epochs to converge to its best results. We plot the results of the first 100 epochs in Figure~\ref{model_complexity}(b) and Figure~\ref{model_complexity}(d) here.
As is shown in Figure~\ref{model_complexity}, when compared with other baseline models, DensE achieves significantly higher performance with the same parameter size or epoch number on both WN18RR and FB15K-237 dataset. 
When comparing the time efficiency, as different models have different training time per epoch, here we also report the training time of DensE and other baseline models. We show that although the more complex math formulation may cause longer training time, the resulting better-designed model can lead to a much faster convergence speed that significantly shortens the total training time under the same machine condition (Table~\ref{time}). 
On the other hand, if we let models have similar performance, e.g., only make DensE reach the final performance of RotatE, we can see it only needs 10 epochs and uses 27\% of RotatE's training time. These results indicate a clear advance of DensE in computation efficiency. We reason that this is mainly achieved by introducing a decoupled scaling operation, thus lowering the embedding dimension required in rotation-only modeling. 


\section{Geometric Interpretation}

\subsection{Theoretical Analysis}

In this section, we first discuss mathematically how DensE provides geometric interpretation of relation patterns including symmetry, antisymmetry, inversion and composition. According to the convention of axis-angle representation described in \textbf{Section 4.2}, all angle-related parameters (\(\theta\), \(\phi\), and \(\psi\)) are restricted to be in the half-closed intervals as we mentioned before. The positive direction of rotation is based on the right-handed coordinate system, the rotation angle with the minus value indicates a rotation opposite to the positive direction. To keep the values of angle to be within the above interval, we relocate angles outside intervals into the desired regions by leveraging the periodic property of the rotation system.   
To begin with, a relation \(r\) is \textbf{symmetric} in DensE if and only if each dimension of its embedding \(\textbf{r}_i\) satisfies \(|\textbf{r}_i|=1\) and the rotation angle satistifies \(\psi_{r_i} = 0 \text{ or } \pi\). 
For \textbf{anti-symmetry} relation pattern, the embedding \(\textbf{r}_i\) satisfies \(|\textbf{r}_i|=1\) , but the rotation angle \(\psi_{r_i}\) should be neither \(0\) nor \(\pi\).
Also, two relations \(r_1\) and \(r_2\) are in \textbf{inverse} pattern, if and only if they satisfy: \(|\textbf{r}_{1i}|*|\textbf{r}_{2i}|=1, \theta_{r_{1i}} = \theta_{r_{2i}}, \phi_{r_{1i}} = \phi_{r_{2i}}\) and \(\psi_{r_{1i}} + \psi_{r_{2i}} = 2\pi\), meaning the embeddings of these two relations share the same rotation axes, but rotate in two opposite directions.

As discussed in \textbf{Section~4.2}, the \textbf{commutative} and \textbf{non-commutative composition} patterns can be naturally modeled by the guarantee of the property of group theory, which covers \textbf{Property~1} of composite relations. Also, following the intuition of \textbf{Property 2}, our model does not enforce a uniform mode of each element in relation representations. Instead, it learns to model the interaction between relations and entities as well as the ambiguity in composition pattern inference, leading to a disperse distribution in the relation embedding space. Last but not least, our model can smoothly deal with constraints posed by relation types in inferring composition patterns, as stated in \textbf{Property 3}. For instance, when modeling the pattern \(r_1(x, y) \Lambda  r_2(y, z) \Rightarrow r_2(x, z) \), the representation from RotatE tends to degenerate to a trivial case where the rotation angle of \(r_1\) and \(r_2\) both set to be \(0 \text{ or } 2\pi \text{ or } r_1=2\pi, r_2=\pi \). In DensE, since the entity embeddings are not required to be perpendicular to the rotation axis, it can also place the embedding of entity \(x\) to be collinear with the rotation axis of \(r_1\). In this way, the rotation axis of \(r_1\) and \(r_2\) are not required to be the same, making the model to more expressive.  In another example, as for the pattern \(r_1(x, y) \Lambda  r_1(y, z) \Rightarrow r_2(x, z) \), besides capturing the relationship of the two rotation angles (i.e., \(\psi_{r_{2i}} = 2\psi_{r_{1i}}\)), the scaling transformation offers an additional degree of freedom, where our model tends to give \(|\textbf{r}_{2i}|=|\textbf{r}_{1i}|^2\). Again, we point out that these ``rules'' are not constant solutions, as the information entities will further guide the model to deviate from the statistical mode for better accommodation of each triplet. Other composition patterns presented in the \textbf{Property 3} can be analyzed in a similar way (see \textbf{Supplementary Note 5.3}).

\subsection{Case Studies}
Here, we show several examples to illustrate the geometric insight given by DensE, which basically reflects the geometric intuition discussed above. 

We start our analysis with \textbf{inverse} relations, which comes from the original WN18 dataset (We have also confirmed the good prediction performance of DensE on the WN18 dataset in~\textbf{Supplementary Note 6}). In Figure~\ref{Fig:Geo}(a), we show the distribution of element-wise addition of embeddings from two \textbf{inverse} relations (\(has\_part\) and \(part\_of\)) of \(\psi\), one representative degree of freedom in modeling relational rotation (Other degrees of freedom can be found in the \textbf{Supplementary Note 5}). In this way, we can visualize how the two embeddings agree with each other. As the two relations are fully inferable by each other, we do observe a clear conjugation as expected in~\textbf{Section 7.1}. This is also reflected in the representation of scaling from these two relations, where the element-wise products tend to be one (Figure~\ref{Fig:Geo}(b)). Interestingly, we do observe that two complementary embeddings are learned these two relations (Figure~\ref{Fig:Geo}(c)-(d)). In particular, the relation whose head entity has a higher semantic hierarchy (i.e., \textit{has\_part}) tend to show a scaling norm \(|Q|\) larger than one, while the relation whose head entity has a lower semantic hierarchy (i.e., \textit{part\_of}) generally shows a scaling norm \(|Q|\) smaller than one. This clearly verifies the intuition of introducing the scaling operation do capture relation-specific semantic hierarchy of entities.

For \textbf{composition patterns}, we slightly change the experiment protocol, with each histogram showing the element-wise difference between the embeddings of a composite relation and the embeddings calculated by multiplying each relation in the relation path. In a case from WN18RR, we demonstrate how DensE models a composition pattern for
\begin{align*}
r_1(h, h') \Lambda r_2(h', t) \Rightarrow r_1(h, t), 
\end{align*}
\noindent where
\begin{gather*}
r_1 = derivationally\_{related}\_{form},\\
r_2 = hypernym.
\end{gather*}
This is a typical case where the composite relation equals the first relation in the relation path. As shown in Figure~2(e), while most embedding dimensions still agree well between the actual composite relation and the calculated relation path, the distribution tends to disperse to a large range, indicating the existence of ambiguity and interaction between entities and relations. To further explore the ambiguity issue in the above case, we perform the same analysis on another small portion of triplets that actually give
\begin{align*}
r_1(h, h') \Lambda r_2(h', t) \Rightarrow r_3(h, t)
\end{align*}
\noindent where
\begin{align*}
r_3 = synset\_domain\_topic\_of.
\end{align*}
Interestingly, we also observe that part of embedding dimensions of \(\mathcal{O}(\textbf{r2})\mathcal{O}(\textbf{r1})\) are aligned with \(\mathcal{O}(\textbf{r3})\) (embedding difference close to zero), demonstrating the flexibility of our model to capture potentially ambiguous relation compositions (Figure~\ref{Fig:Geo}(f)). On the other hand, the model can also learn to put the rotation axis of \(r_1\) collinear with the embedding of head entities \(h\) in the composition mode expressed as
\begin{align*}
r_1(h, h') \Lambda r_2(h', t) \Rightarrow r_1(h, t)
\end{align*}
\noindent reflecting the interaction of entities and relations (Figure~\ref{Fig:Geo}(g)). This example clearly demonstrates the interpretability of DensE in modeling complex composition patterns. The geometric patterns for other relation patterns can be found in \textbf{Supplementary Note 5}. Together with the properties discussed in \textbf{Section~3}, here we clearly demonstrate the pros and cons of the current rotation-based translational KGE method in modeling composition relation patterns.

\section{Conclusion}
In this work, we propose an effective method, named DensE, for knowledge graph embedding. DensE decomposes a relation operator into an SO(3) group-based rotation as well as a scaling transformation. Extensive experiments show that DensE possesses good performance in knowledge completion with high computational efficiency. Also, DensE provides a straightforward geometric interpretation for the relations, leading to meaningful insights for the future work for modeling complex relation patterns.

\section*{Declaration of competing interest}
The authors declare that they have no known competing financial interests or personal relationships that could have appeared to influence the work reported in this paper.

\section*{Acknowledgements}
The authors thank Dr. Y. Wen, Dr. W. Peng, Dr. D. Wang, Dr. W. Guo, Mr. A. Shen and Ms. X. Lin for insightful comments on the manuscript. We also thank Dr. Y. Guo and Ms. C. Jiang for helpful suggestions in the experimental settings. We also thank all the colleagues in AI Application Research Center (AARC) of Huawei Technologies for their supports.


\appendix

\section{The geometrical interpretation of non-commutative compositions}

\begin{figure}[!ht]
\centering     
\includegraphics[width=90mm]{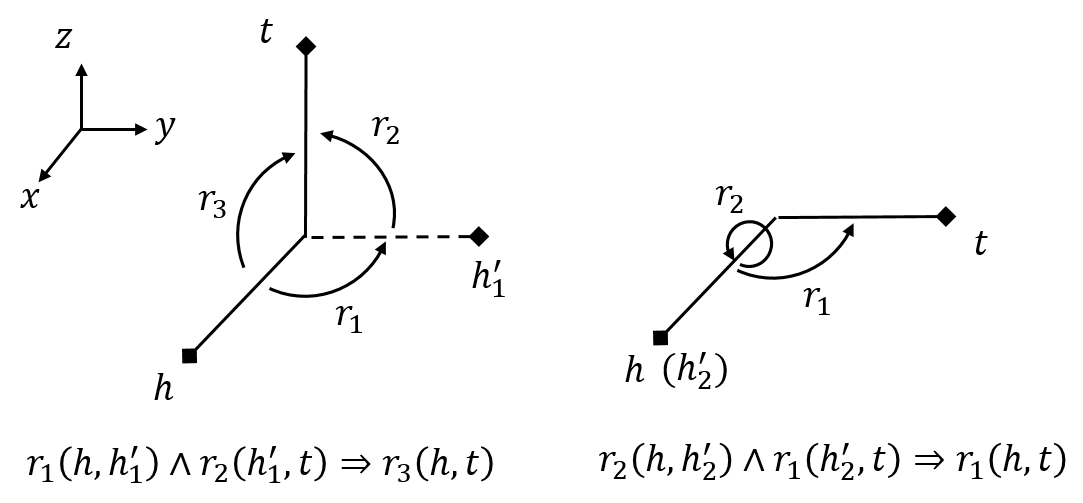}
\caption{A simple example for how rotation in the 3-D Euclidean space can model non-commutative relations}
\label{simple example for how rotation in 3D-Euclidean space can model non-commutative relations}
\end{figure}

\textbf{Left}: A rotation about axis-\(z\) followed by a rotation about axis-\(x\), the initial vector \(\textbf{h}\) is placed along axis-\(x\). It can be seen that the final state is along axis-\(z\), and two rotation operations are equivalent to one operation with the rotation axis to be about axis-\(y\).

\textbf{Right}: The rotation operation sequence is reversed from the left figure. A rotation about axis-\(x\) is followed by a rotation about axis-\(z\). The final state is then changed to be along axis-\(y\). Since the initial vector is collinear with the first rotation axis, the two rotation operations are equal to the last rotation (rotation about axis-\(z\)). 

\section{Hyperparameters setting}
The ranges of the hyperparameters for the grid search are set as follows: Embedding size \(k \in \{100, 200, 500, 1000\}\) (In our model, each entity is represented with a matrix with a size of \(3 \times k\), and each relation with a matrix with a size of \(4 \times k\)), batch size \(b \in \{256, 512, 1024\}\), fixed margin \(\gamma \in \{3.0, 6.0, 9.0, 12.0, 15.0, 24.0, 30.0\}\), negative sampling size \(n \in \{256, 512, 1024\}\), self-adversarial sampling temperature  \(\alpha \in \{0.3, 0.5, 1.0\}\). The initial learning rate \(\eta\) is set to be \(0.1\), and it decays with a factor of \(1/2\) if the training loss does not decrease in \(1000\) epochs. We list the best hyperparameters setting of DensE on the benchmark datasets in Supplementary Table~\ref{HYP}.
\begin{table*}[h!]
\centering
\caption{Hyperparameters settings of DensE in this study.}

 \begin{tabular}{ l c c c c c  }
  \toprule
  Dataset  & \Centerstack{Embedding \\ size \(k\)} & \Centerstack{Batch \\ size \(b\)}  & \Centerstack{Margin \\ \(\gamma\)}  & \Centerstack{Negative \\sample size \(n\)}  & \Centerstack{adv temperature \\ \(\alpha\)} \\  
  \midrule
  WN18  & 200    & 512  & 12.0   & 1024   & 0.3 \\ 
  WN18RR & 300   & 512  & 6.0  & 512  & 0.5   \\
  FB15k-237 & 800   & 1024  & 9.0   & 256   & 1.0\\ 
  YAGO3-10    & 200    & 1024  & 24.0   & 512   & 1.0 \\ 
 \bottomrule
\end{tabular} 
\label{HYP}
\end{table*}



\section{Additional performance metrics}

For a more complete comparison of each method, for each dataset we list MR, MRR, H@1, H@3, and H@10 in Supplementary Tables~\ref{supp results table wn18rr},~\ref{supp results table fb15k237}, and~\ref{supp results table yago}.

\section{Effect of self-adversarial negative sampling on DensE and RotatE}
In the ablation study, we observe a significant contribution of the self-adversarial negative sampling technique on the prediction performance of FB15k-237 and YAGO3-10. Therefore, we compare our model with RotatE in the setting where both models are trained without self-adversarial negative sampling (Supplementary Table~\ref{without adv}). These results further confirm the superiority of our model without self-adversarial negative sampling.

\begin{table*}[!h]
\centering
\caption{Performance comparison on WN18RR. Best results are labeled in bold and the second best are underlined. \textit{First} indicates the originally reported performance of each method. \textit{Recent} shows the best results of more selected recent models. \textit{Ours} reports the performance of DensE. For MRR, the lower, the better; for other metrics, the higher, the better. }

\begin{tabular}{ccccccc}
  \toprule
    \multicolumn{2}{c}{} & \multicolumn{5}{c}{\textbf{WN18RR}}  \\
   	~   &   Model   &   MR &   MRR  &   H@1 &   H@3 &   H@10    \\
    \midrule
    \multirow{4}*{\textit{First}}	&	RESCAL~\cite{wang2018evaluating}	&	-	&	0.420	&	-	&	-	&	0.447	\\
    ~	&	TransE~\cite{nguyen2017novel}	&	3384	&	0.226	&	-	&	-	&	0.501	\\
    ~	&	DistMult~\cite{dettmers2018convolutional}	&	5110	&	0.430	&	0.390	&	0.440	&	0.490	\\
    ~	&	ComplEx~\cite{dettmers2018convolutional}	&	5261	&	0.440	&	0.410	&	0.460	&	0.510	\\
    \midrule													
    \multirow{6}*{\textit{Recent}} 	&	RotatE~\cite{sun2019rotate}	&	3340	&	0.476	&	0.428	&	0.492	&	0.571	\\
    ~	&	NagE~\cite{Yang2020AGF}	&	-	&	0.477	&	0.432	&	0.493	&	0.574	\\
    ~	&	QuatE~\cite{jia2019quaternions}	&	3472	&	0.481	&	0.436	&	0.500	&	0.564	\\
    ~	&	D4-STE~\cite{xu2019relation} 	&	-	&	0.480	&	\textbf{0.452}	&	0.491	&	0.536	\\
    ~	&	TuckER~\cite{balavzevic2019tucker}	&	-	&	0.470	&	\underline{0.443}	&	0.482	&	0.526	\\
    ~	&	Rotate3D~\cite{gao2020rotate3d}	&	\underline{3328}	&	0.489	&	0.442	&	0.505	&	0.579	\\ 
    ~	&	HAKE~\cite{zhang2020learning}	&	-	&	\textbf{0.497}	&	\textbf{0.452}	&	\textbf{0.516}	&	\underline{0.582}	\\
    \midrule													
    \textit{Ours}	&	DensE	&	\textbf{2934}	&	\underline{0.492}	&	\underline{0.443}	&	\underline{0.509}	&	\textbf{0.586}	\\

\bottomrule
\end{tabular} 
\label{supp results table wn18rr}
\end{table*}

\begin{table*}[!h]
\centering
\caption{Performance comparison on FB15K-237. Best results are labeled in bold and the second best are underlined. \textit{First} indicates the originally reported performance of each method. \textit{Recent} shows the best results of more selected recent models. \textit{Ours} reports the performance of DensE. For MRR, the lower, the better; for other metrics, the higher, the better. }

\begin{tabular}{ccccccc}
  \toprule
    \multicolumn{2}{c}{} & \multicolumn{5}{c}{\textbf{FB15K-237}}  \\
   	~   &   Model   &   MR &   MRR  &   H@1 &   H@3 &   H@10    \\
    \midrule
    \multirow{4}*{\textit{First}}	&	RESCAL~\cite{wang2018evaluating}	&	 -	&	0.270	&	-	&	 -	&	0.427	\\
    ~	&	TransE~\cite{nguyen2017novel}	&	357	&	0.294	&	-	&	-	&	0.465	\\
    ~	&	DistMult~\cite{dettmers2018convolutional}	&	254	&	0.241	&	0.155	&	0.263	&	0.419	\\
    ~	&	ComplEx~\cite{dettmers2018convolutional}	&	339	&	0.247	&	0.158	&	0.275	&	0.428	\\
    \midrule													
    \multirow{6}*{\textit{Recent}} 	&	RotatE~\cite{sun2019rotate}	&	177	&	0.338	&	0.241	&	0.375	&	0.533	\\
    ~	&	NagE~\cite{Yang2020AGF}	&	-	&	0.340	&	0.244	&	0.378	&	0.530	\\
    ~	&	QuatE~\cite{jia2019quaternions}	&	176	&	0.311	&	0.221	&	0.342	&	0.495	\\
    ~	&	D4-STE~\cite{xu2019relation} 	&	-	&	0.320	&	0.230	&	0.353	&	0.502	\\
    ~	&	TuckER~\cite{balavzevic2019tucker}	&	-	&	\textbf{0.358}	&	\textbf{0.266}	&	\textbf{0.394}	&	\textbf{0.544}	\\
    ~	&	Rotate3D~\cite{gao2020rotate3d}	&	\underline{165}	&	0.347	&	0.250	&	0.385	&	\underline{0.543} 	\\ 
    ~	&	HAKE~\cite{zhang2020learning}	&	-	&	0.346	&	0.250	&	0.381	&	0.542	\\
    \midrule													
    \textit{Ours}	&	DensE	&	\textbf{161}	&	\underline{0.351}	&	\underline{0.256}	&	\underline{0.386}	&	\textbf{0.544}	\\

\bottomrule
\end{tabular} 
\label{supp results table fb15k237}
\end{table*}

\begin{table*}[!h]
\centering
\caption{Performance comparison on YAGO3-10. Best results are labeled in bold and the second best are underlined. \textit{First} indicates the originally reported performance of each method. \textit{Recent} shows the best results of more selected recent models. \textit{Ours} reports the performance of DensE. For MRR, the lower, the better; for other metrics, the higher, the better. }

\begin{tabular}{ccccccc}
  \toprule
    \multicolumn{2}{c}{} & \multicolumn{5}{c}{\textbf{YAGO3-10}}  \\
   	~   &   Model   &   MR &   MRR  &   H@1 &   H@3 &   H@10    \\
    \midrule
    \multirow{2}*{\textit{First}}   	& 	DistMult~\cite{dettmers2018convolutional}	&	5926	&	0.340	&	0.240	&	0.380	&	0.540	\\
    ~	&	ComplEx~\cite{dettmers2018convolutional}	&	6351	&	0.360	&	0.260	&	0.400	&	0.550	\\
    \midrule													
    \multirow{3}*{\textit{Recent}} 	&	RotatE~\cite{sun2019rotate}	&	\underline{1767}	&	0.495	&	0.402	&	0.550	&	0.670	\\
    ~	&	D4-STE~\cite{xu2019relation} 	&	-	&	0.472	&	0.381	&	0.523	&	0.643	\\
    ~	&	HAKE~\cite{zhang2020learning}	&	-	&	\textbf{0.545}	&	\underline{0.462}	&	\textbf{0.596}	&	\textbf{0.694}	\\
    \midrule													
    \textit{Ours}	&	DensE	&	\textbf{1450}	&	\underline{0.541}	&	\textbf{0.465}	&	\underline{0.585}	&	\underline{0.678}	\\

\bottomrule
\end{tabular} 
\label{supp results table yago}
\end{table*}

\begin{table*}[!h]
\centering
\caption{Results of DensE and RotatE without self-adversarial negative sampling training technique, where “adv” represents “self-adversarial”.}

 \begin{tabular}{ l c c c c  }
  \toprule
  Model & WN18 & WN18RR & FB15k-237  & YAGO3-10 \\  
  \midrule
  DensE (w/o adv)  & 0.950    & 0.486  & 0.306   & 0.452  \\ 
  RotatE (w/o adv)  & 0.947    & 0.470  & 0.297   & 0.439 \\ 
 \bottomrule
\end{tabular} 
\label{without adv}
\end{table*}

\section{Capability of DensE in modeling relation patterns}


In this Section, we provide a detailed analysis on how our method tend to model each relation pattern in an interpretable way. In our experiment, we calculate the statistical rule of each degree of freedom to reflect the effect of specific relation patterns. In addition, we also sometimes compare the embeddings of two relation types (or a relation type and an entity)~\textit{per element}, i.e., we perform element-wise addition, subtraction, multiplication on each embedding dimension. Then, we use the distribution of these results to demonstrate how the two compared embeddings agree with each other. Note that below we use an addition subscript \(i\) to denote each dimension in the embeddings.


\subsection{Symmetry/anti-symmetry pattern}
\begin{figure*}[htbp]
\centering
\begin{minipage}{6cm}
\centering
\includegraphics[width=4.8cm]{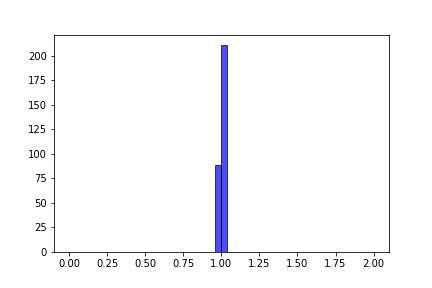}
\subcaption{\tiny{\(|Q_{derivationally\_related\_form}|\)}}
\end{minipage}
\begin{minipage}{6cm}
\centering
\includegraphics[width=4.8cm]{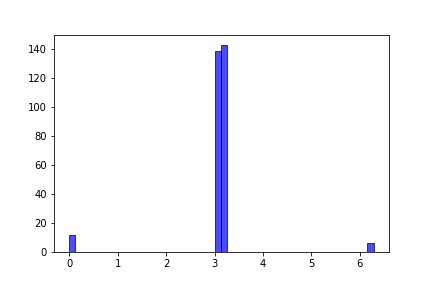}
\subcaption{\tiny{\(\psi_{derivationally\_related\_form}\)}}
\end{minipage}

\begin{minipage}{6cm}
\centering
\includegraphics[width=4.8cm]{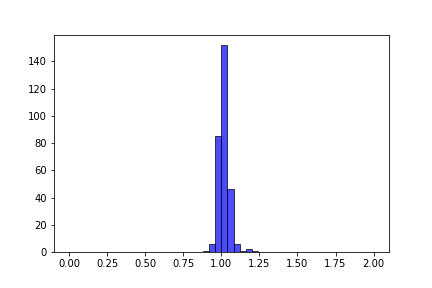}
\subcaption{\tiny{\(|Q_{also\_see}|\)}}
\end{minipage}%
\begin{minipage}{6cm}
\centering
\includegraphics[width=4.8cm]{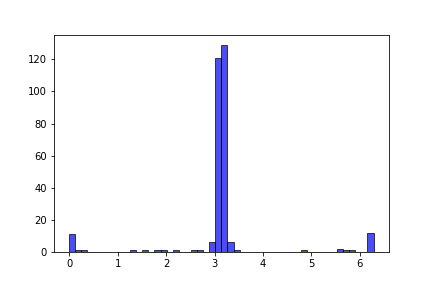}
\subcaption{\tiny{\(\psi_{also\_see}\)}}
\end{minipage}

\begin{minipage}{6cm}
\centering
\includegraphics[width=4.8cm]{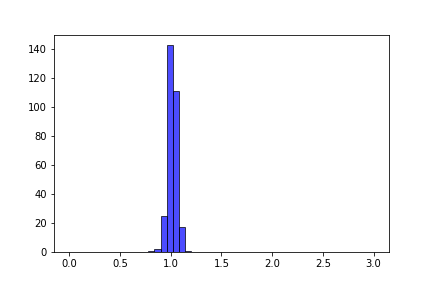}
\subcaption{\tiny{\(|Q_{similar\_to}|\)}}
\end{minipage}%
\begin{minipage}{6cm}
\centering
\includegraphics[width=4.8cm]{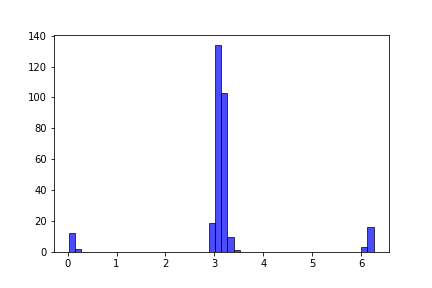}
\subcaption{\tiny{\(\psi_{similar\_to}\)}}
\end{minipage}

\begin{minipage}{6cm}
\centering
\includegraphics[width=4.8cm]{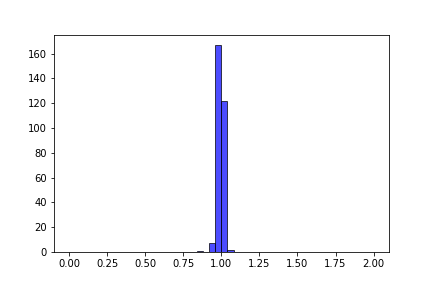}
\subcaption{\tiny{\(|Q_{verb\_group}|\)}}
\end{minipage}%
\begin{minipage}{6cm}
\centering
\includegraphics[width=4.8cm]{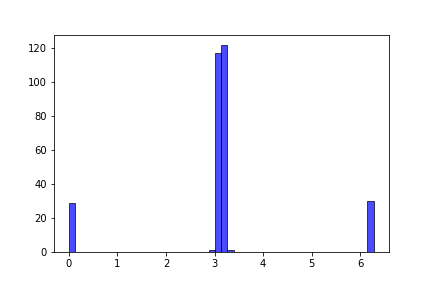}
\subcaption{\tiny{\(\psi_{verb\_group}\)}}
\end{minipage}

\begin{minipage}{6cm}
\centering
\includegraphics[width=4.8cm]{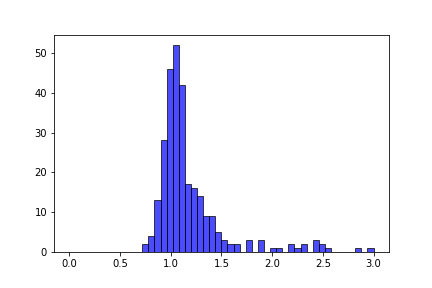}
\subcaption{\tiny{\(|Q_{member\_meronym}|\)}}
\end{minipage}%
\begin{minipage}{6cm}
\centering
\includegraphics[width=4.8cm]{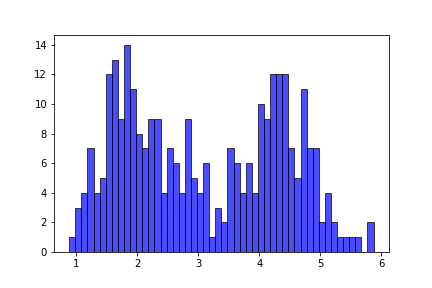}
\subcaption{\tiny{\(\psi_{member\_meronym}\)}}
\end{minipage}
\caption{Geometric interpretation of how DensE models symmetry patterns and anti-symmetry patterns. Each row shows the distribution of \(|Q|\) and \(\psi\) for a given relation, respectively.}
\label{symmetric plots}
\end{figure*}

As pointed out in main text \textbf{Section 7.1}, for the \textbf{symmetry} relation pattern, the scaling factor \(|Q|\) of symmetric relation tend to be one, and the rotation angle \(\psi\) should be \(0\) or \(\pi\) in \([0, 2\pi)\). For \textbf{anti-symmetry} relation pattern, one can easily check that the scaling factor \(|Q|\) should also be one, but the rotation angle \(\psi\) should be neither \(0\) nor \(\pi\) in the range of \([0, 2\pi)\). Here we show the distributions of rotation angle \(\psi\) and scaling factor \(|Q|\) of four relations with \textbf{symmetry} pattern in WN18RR (Supplementary Figure~\ref{symmetric plots} (a)-(h)). We also show distributions of anti-symmetry relation pattern ``member\_meronym'' in Supplementary Figure~\ref{symmetric plots} (i)-(j). As we can see that the scaling factor is roughly around one. For the rotation angle \(\psi\), there are just few elements fall into the bin that contains \(\pi\). 
It should be noted that since the embedding size \(k\) in our model for WN18RR is set to be 300, the sum of frequency in these distributions also equals to 300. 

\subsection{Inversion pattern}
In main text \textbf{Section 7.1}, we assert that if two relations \(r_1\) and \(r_2\) satisfy the \textbf{inverse} pattern, if and only if they satisfy: \(|\textbf{r}_{1i}|*|\textbf{r}_{2i}|=1, \theta_{r_{1i}} = \theta_{r_{2i}}, \phi_{r_{1i}} = \phi_{r_{2i}}\) and \(\psi_{r_{1i}} + \psi_{r_{2i}} = 2\pi\). In Supplementary Figure~\ref{inverse plots1}, we show a case of paired relations with \textbf{inversion} pattern from WN18 dataset, namely \(has\_part, part\_of\). We plot the scaling factor \(|Q|\), magnitude of the rotation \(\psi\), and \((\theta, \phi)\) that describe the rotation axis for each relation (first two columns), as well as their element-wise alignment results (the last column).


\begin{figure*}[htbp]
\centering
\begin{minipage}{3.9cm}
\centering
\includegraphics[width=3.9cm]{images/wn18_has_part_r.png}
\subcaption{\tiny{\(|Q_{has\_part}|\)}}
\end{minipage}
\begin{minipage}{3.9cm}
\centering
\includegraphics[width=3.9cm]{images/wn18_part_of_r.png}
\subcaption{\tiny{\(|Q_{part\_of}|\)}}
\end{minipage}
\begin{minipage}{3.9cm}
\centering
\includegraphics[width=3.9cm]{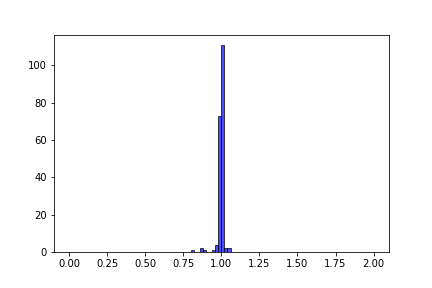}
\subcaption{\tiny{\(|Q_{has\_part}| * |Q_{part\_of}|\)}}
\end{minipage}

\begin{minipage}{4cm}
\centering
\includegraphics[width=4cm]{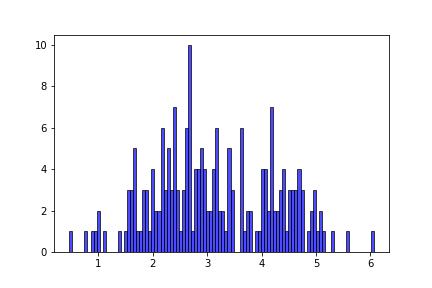}
\subcaption{\tiny{\(\psi_{has\_part}\)}}
\end{minipage}%
\begin{minipage}{4cm}
\centering
\includegraphics[width=4cm]{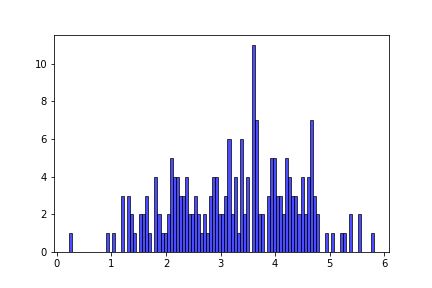}
\subcaption{\tiny{\(\psi_{part\_of}\)}}
\end{minipage}
\begin{minipage}{4cm}
\centering
\includegraphics[width=4cm]{images/wn18_has_part_part_of_inverse_psi.png}
\subcaption{\tiny{\(\psi_{has\_part} + \psi_{part\_of}\)}}
\end{minipage}

\begin{minipage}{4cm}
\centering
\includegraphics[width=4cm]{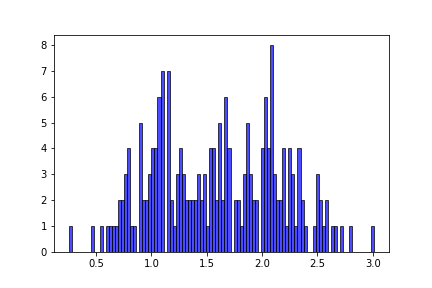}
\subcaption{\tiny{\(\theta_{has\_part}\)}}
\end{minipage}%
\begin{minipage}{4cm}
\centering
\includegraphics[width=4cm]{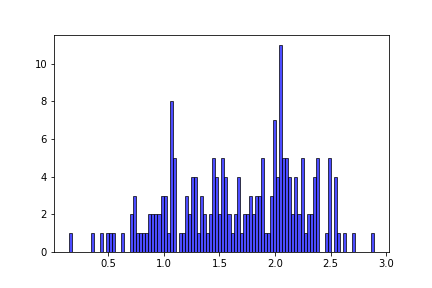}
\subcaption{\tiny{\(\theta_{part\_of}\)}}
\end{minipage}
\begin{minipage}{4cm}
\centering
\includegraphics[width=4cm]{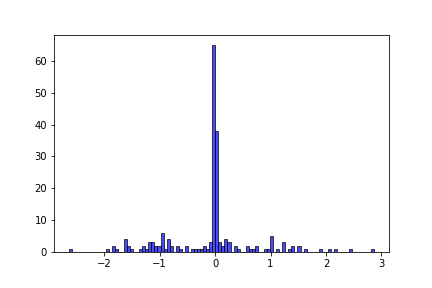}
\subcaption{\tiny{\(\theta_{has\_part} - \theta_{part\_of}\)}}
\end{minipage}

\begin{minipage}{4cm}
\centering
\includegraphics[width=4cm]{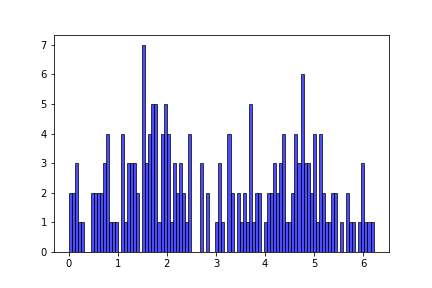}
\subcaption{\tiny{\(\phi_{has\_part}\)}}
\end{minipage}%
\begin{minipage}{4cm}
\centering
\includegraphics[width=4cm]{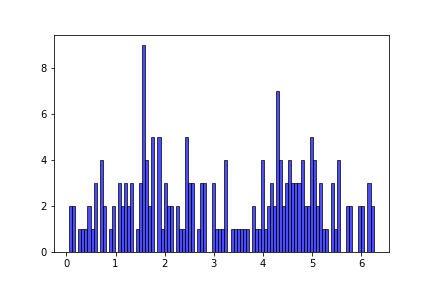}
\subcaption{\tiny{\(\phi_{part\_of}\)}}
\end{minipage}
\begin{minipage}{4cm}
\centering
\includegraphics[width=4cm]{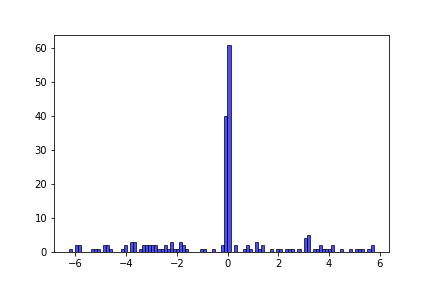}
\subcaption{\tiny{\(\phi_{has\_part} - \phi_{part\_of}\)}}
\end{minipage}

\caption{Geometric interpretation of how DensE models inverse pattern, using an example of \((has\_part, part\_of)\) from WN18. At each row, we show the embedding from one degree of freedom of our model. The first two columns show the embeddings of each relation type, and the last column shows the alignment of the two embeddings regarding a specific degree of freedom. }
\label{inverse plots1}
\end{figure*}

\subsection{Composition pattern}

\subsubsection{Ambiguity in composition pattern}

In a real-world KG (here we take a sub-graph from WN18RR as an example), due to the ambiguity issue mentioned in the main text (composition pattern~\textbf{Property 2}), there exist plenty of examples where a third relation (the composite relation) cannot be inferred given the two participating relations alone (Supplementary Figure~\ref{WN18rr instance}(a)). For example, given \(r_1 = derivationally\_{related}\_{form}\) and \( r_2 = hypernym\), we have the composition pattern as shown with the blue lines: (Trade(VB), derivationally\_{related}\_{form}, Trade(NN)), (Trade(NN), hypernym, transaction) and (Trade(VB), derivationally\_{related}\_{form} , transaction). From these cases, it seems that one can summarize the composition pattern as: \(r_3 = r_1 = derivationally\_{related}\_{form}\), i.e., \(r_1(h, h') \Lambda r_2(h', t) \Rightarrow r_1(h, t) \). However, we also have the triangle with red lines, i.e., (Trade(VB), derivationally\_{related}\_{form}, Selling), (Selling, hypernym, mercantilism) and (Trade(VB), synset\_domain\_topic\_of, mercantilism). In these cases, it looks like the composition pattern has the form that \(r_1(h, h') \Lambda r_2(h', t) \Rightarrow r_3(h, t) \), where \(r_3 = synset\_domain\_topic\_of\). This ambiguity means that the composition mode is not uniform but depends on specific entities and their other neighborhoods. Therefore, in order to give the model sufficient flexibility to learn this, our model does not require all the dimensions in a relation embedding to fit in one single composition mode (e.g., \(r_1(h, h') \Lambda r_2(h', t) \Rightarrow r_1(h, t) \) or \(r_1(h, h') \Lambda r_2(h', t) \Rightarrow r_3(h, t) \)). In consequence, the learned relation embedding for a composite relation are actually distributed in a disperse manner, with the majority of embedding dimensions following mode \(r_1(h, h') \Lambda r_2(h', t) \Rightarrow r_1(h, t) \), and some minor portions following \(r_1(h, h') \Lambda r_2(h', t) \Rightarrow r_3(h, t) \), which is consistent with the abundance of each mode in the training data.

\subsubsection{Case study: a two-hop composition pattern}

As we mentioned in the main text, in a composition pattern, the relations involved are not necessarily different (composition pattern~\textbf{Property 3}). Here we discuss the simplest case of a two-hop relation composition. Even under the assumption that the composite relation can be inferred from its composition alone (i.e., an \textbf{unambiguous} composition), there are still five possible situations, as shown in Supplementary Figure~\ref{WN18rr instance}(b).


\begin{figure*}[htbp]
\centering
\begin{minipage}{10cm}
\centering
\includegraphics[width=10cm]{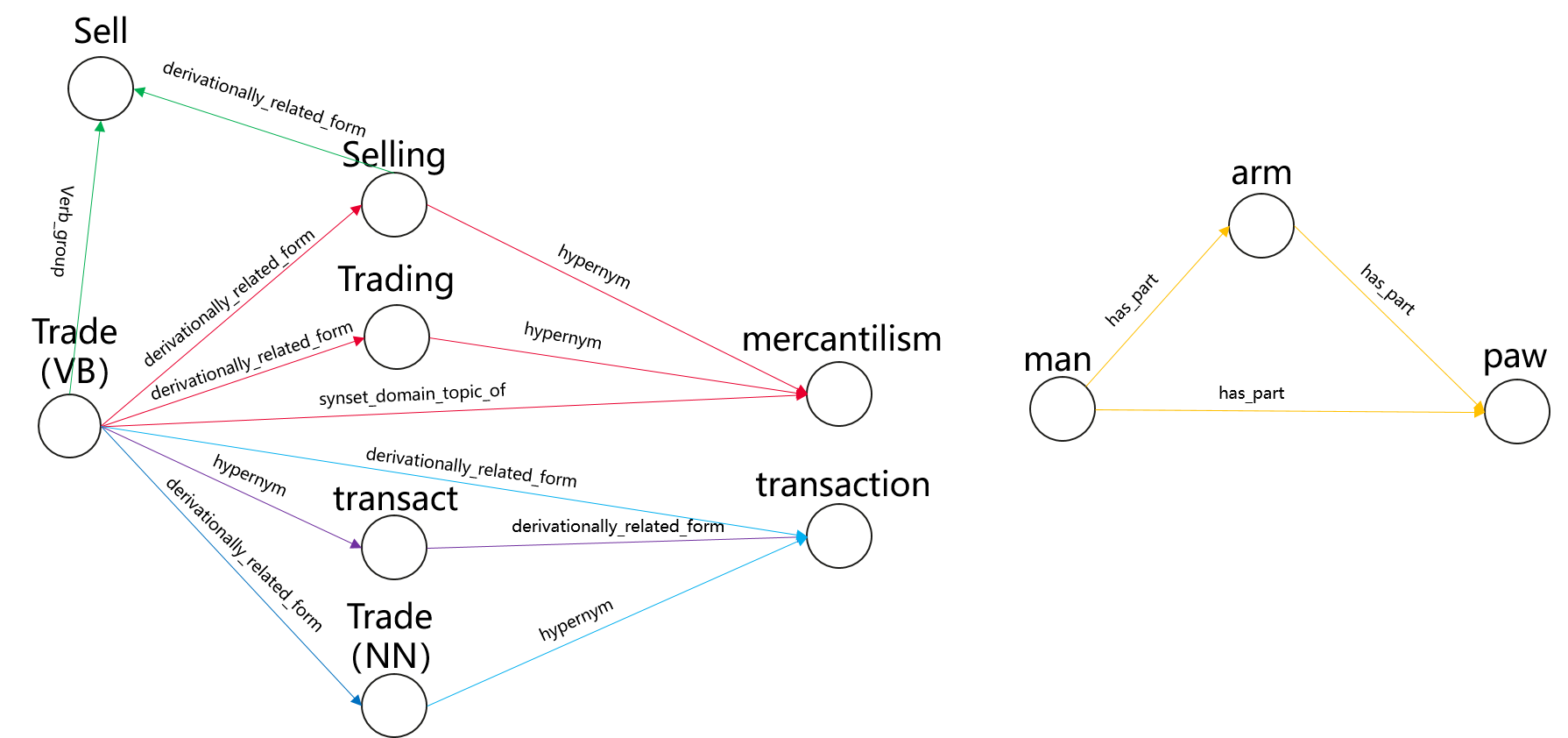}
\subcaption{}
\end{minipage}

\begin{minipage}{8cm}
\centering
\includegraphics[width=8cm]{images/1c.png}
\subcaption{}
\end{minipage}
\caption{(a) A representative subgraph from WN18RR to show ambiguity. (b) The relations involved in a composition pattern are not necessarily to be mutually different.}
\label{WN18rr instance}
\end{figure*}


Here we provide a detailed analysis on these five situations, using real cases in WN18RR as examples (Supplementary Figure~\ref{WN18rr instance}(a)).

\textbf{Situation 1:} \(r_1\), \(r_2\) and \(r_3\) are all the same, e.g., the triangle in yellow color with three triplets: (\(man\), \(has\_part\), \(arm)\), (\(arm\), \(has\_part\), \(paw)\) and (\(man\), \(has\_part\), \(paw\)). To satisfy the composition relation constraint, in principle the model can adopt several approaches to fit this situation. From the perspective of relation embedding, it can learn to fit \(|Q_{has\_part}|=1\) or \(\psi_{has\_part} \in \{0, 2\pi\}\). Also, in terms of interaction between entities and relations, it can also learn to align the rotation axis of relation with the entity embedding (see Supplementary Figure~\ref{has part}(c), showing the differences between the \(\theta\) component of \(r_1\) in the mode \(r_1(h, h') \Lambda r_1(h', t) \Rightarrow r_1(h, t) \) and the corresponding head entities \(h\) in spherical coordinate system). Note that when the rotation axis aligns with head entity embedding in spherical coordinate system, the composition pattern can be modeled by the scaling factor alone, i.e., \(|Q_{has\_part}|^2 = |Q_{has\_part}|\), regardless of the rotation magnitude \(\psi\). 

In this case, we see the model mainly adopts the first approach (Supplementary Figure~\ref{has part}(a)), as \(|Q_{has\_part}|\) is roughly around 1. In addition, the partial alignment between relation rotation axis and entities is also observed (Supplementary Figure~\ref{has part}(c)). In contrast, for \(\psi_{has\_part}\), we see very few dimensions satisfy the condition (\(0 \text{ or } 2\pi\)), which is also consistent with the dominate role of scaling factor in this case. Noticed the fact that the relation ``hypernym'' dominates the composite patterns (Refer to Table 3 in the main text), we also find that \(r_1=r_2=r_3=\text{hypernym}\) pattern exists in the WN18RR dataset. The above similar analysis can be done for this case. The relevant results are shown in Supplementary Figure~\ref{has part}(d), (e), and (f), respectively.



\begin{figure*}[htbp]
\centering
\begin{minipage}{3.8cm}
\centering
\includegraphics[width=3.8cm]{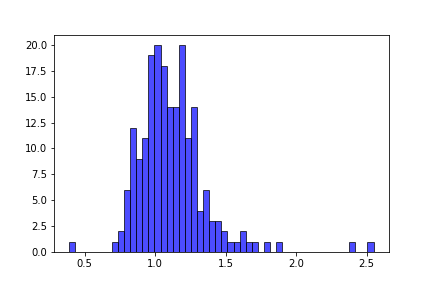}
\subcaption{\tiny{\(|Q_{has\_part}|\)}}
\end{minipage}
\begin{minipage}{3.8cm}
\centering
\includegraphics[width=3.8cm]{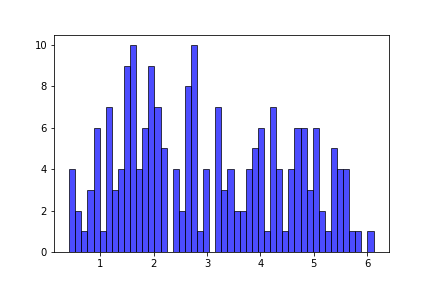}
\subcaption{\tiny{\(\psi_{has\_part}\)}}
\end{minipage}
\begin{minipage}{3.8cm}
\centering
\includegraphics[width=3.8cm]{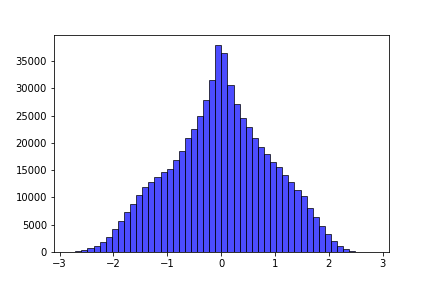}
\subcaption{\tiny{\(\theta(\mathcal{O}(\textbf{r1})) - \theta(\textbf{h})\) (has\_part)}}
\end{minipage}
\begin{minipage}{3.8cm}
\centering
\includegraphics[width=3.8cm]{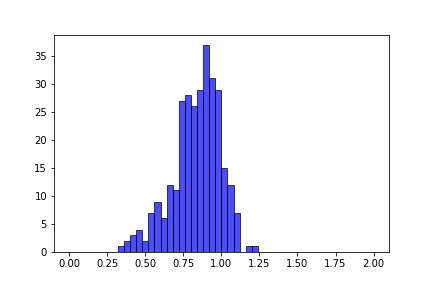}
\subcaption{\tiny{\(|Q_{hypernym}|\)}}
\end{minipage}
\begin{minipage}{3.8cm}
\centering
\includegraphics[width=3.8cm]{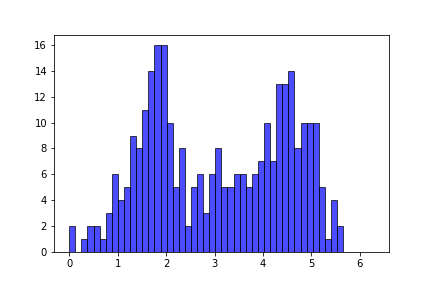}
\subcaption{\tiny{\(\psi_{hypernym}\)}}
\end{minipage}
\begin{minipage}{3.8cm}
\centering
\includegraphics[width=3.8cm]{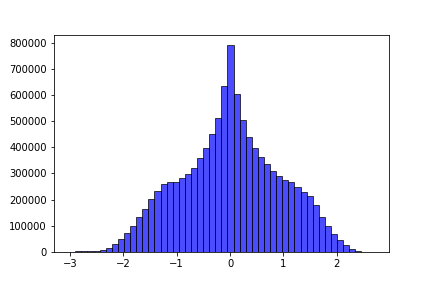}
\subcaption{\tiny{\(\theta(\mathcal{O}(\textbf{r1})) - \theta(\textbf{h})\) (hypernym)}}
\end{minipage}
\caption{Geometric interpretation of composition patterns in \textbf{Situation 1}.}
\label{has part}
\end{figure*}

\textbf{Situation 2:} \(r_1\) and \(r_2\) are the same, but not equal to \(r_3\). An example is given by the triangle in green with three triplets: (\(Trade(VB)\), \(derivationally\_related\_form\), \(Selling\)), (\(Selling\), \(derivationally\_related\_form\), \(Sell\)) and (\(Trade(VB)\), \(verb\_group\), \(Sell\)). As mentioned in main text \textbf{Section 7.1}, we expect the model to learn \(\psi_{verb\_group} = 2\psi_{derivationally\_related\_form}\) and \(|Q_{verb\_group}|=|Q_{derivationally\_related\_form}|^2\), both of which are confirmed in this case. Besides this, we also observe that the rotation axes of the two relations are aligned on some dimensions, i.e, \(\theta_{verb\_group} = \theta_{derivationally\_related\_form}\) and \(\phi_{verb\_group} = \phi_{derivationally\_related\_form}\). The corresponding distributions are shown in Supplementary Figure~\ref{119}, where we have \(r_1=r_2=derivationally\_related\_form\) and \(r_3=verb\_group\). 

\begin{figure}[htbp]
\centering
\begin{minipage}{3.5cm}
\centering
\includegraphics[width=3.5cm]{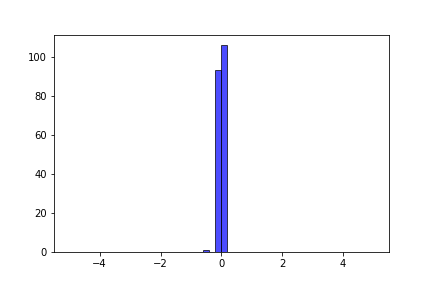}
\subcaption{\tiny{\(Q(\mathcal{O}(\textbf{r3})-Q^{2}(\mathcal{O}(\textbf{r1}))\)}}
\end{minipage}
\begin{minipage}{3.5cm}
\centering
\includegraphics[width=3.5cm]{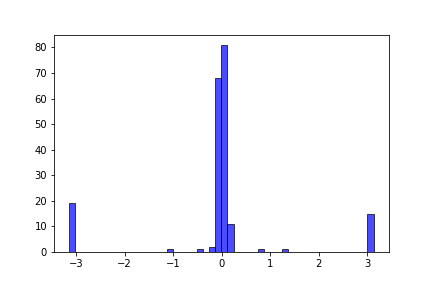}
\subcaption{\tiny{\(\psi(\mathcal{O}(\textbf{r3}))-2*\psi(\mathcal{O}(\textbf{r1}))\)}}
\end{minipage}
\\
\begin{minipage}{3.5cm}
\centering
\includegraphics[width=3.5cm]{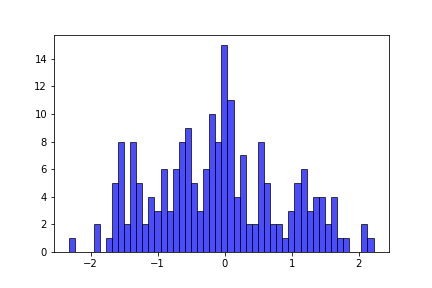}
\subcaption{\tiny{\(\theta(\mathcal{O}(\textbf{r3}))-\theta(\mathcal{O}(\textbf{r1}))\)}}
\end{minipage}
\begin{minipage}{3.5cm}
\centering
\includegraphics[width=3.5cm]{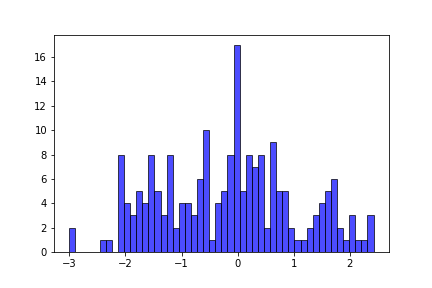}
\subcaption{\tiny{\(\phi(\mathcal{O}(\textbf{r3}))-\phi(\mathcal{O}(\textbf{r1}))\)}}
\end{minipage}
\caption{Geometric interpretation of composition patterns in \textbf{Situation 2}.}
\label{119}
\end{figure}

\textbf{Situation 3:} \(r_1\) and \(r_3\) are the same, but not equal to \(r_2\), e.g., the triangle in blue color with three triplets set: (\(Trade(VB)\),\(derivationally\_related\_form\)
, \(Trade(NN)\)), (\(Trade(NN)\), \(hypernym\), \(transaction\)) and (\(Trade(VB)\), \(derivationally\_related\_form\), \(transaction\)). Here we compare the difference between the embedding of a composite relation and the embedding calculated by multiplying each relation in the relation path. We have already shown the distribution of \(\psi\) in the main context. Other distributions are shown in Supplementary Figure~\ref{101}, where we have \(r_1=r_3=derivationally\_related\_form\) and \(r_2=hypernym\).

\begin{figure}[htbp]
\centering
\begin{minipage}{3.5cm}
\centering
\includegraphics[width=3.5cm]{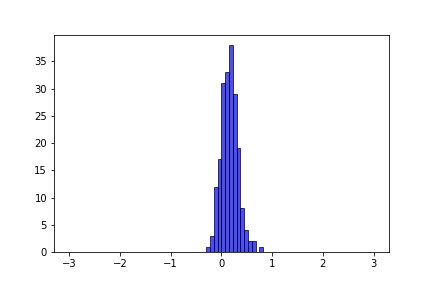}
\subcaption{\tiny{\(Q(\mathcal{O}(\textbf{r2})\mathcal{O}(\textbf{r1}))-Q(\mathcal{O}(\textbf{r1}))\)}}
\end{minipage}
\begin{minipage}{3.5cm}
\centering
\includegraphics[width=3.5cm]{images/wn18rr_derivationally_related_form_hypernym_derivationally_related_form_composition_theta_r.png}
\subcaption{\tiny{\(\psi(\mathcal{O}(\textbf{r2})\mathcal{O}(\textbf{r1}))-\psi(\mathcal{O}(\textbf{r1}))\)}}
\end{minipage}
\\
\begin{minipage}{3.5cm}
\centering
\includegraphics[width=3.5cm]{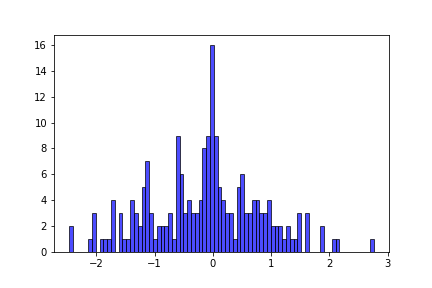}
\subcaption{\tiny{\(\theta(\mathcal{O}(\textbf{r2})\mathcal{O}(\textbf{r1}))-\theta(\mathcal{O}(\textbf{r1}))\)}}
\end{minipage}
\begin{minipage}{3.5cm}
\centering
\includegraphics[width=3.5cm]{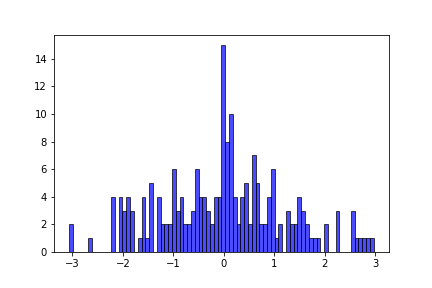}
\subcaption{\tiny{\(\phi(\mathcal{O}(\textbf{r2})\mathcal{O}(\textbf{r1}))-\phi(\mathcal{O}(\textbf{r1}))\)}}
\end{minipage}
\caption{Geometric interpretation of composition patterns in \textbf{Situation 3}.}
\label{101}
\end{figure}

\textbf{Situation 4:} \(r_2\) and \(r_3\) are the same, but not equal to \(r_1\), e.g., the triangle in purple color with three triplets set: (\(Trade(VB)\), \(hypernym\), \(transact\)), (\(transact\), \(derivationally\_related\_form\), \(transaction\)) and (\(Trade(VB)\), \(derivationally\_related\_form\), \(transaction\)). Here we compare the difference between the embedding of a composite relation and the embedding calculated by multiplying each relation in the relation path. Related distributions are shown in Supplementary Figure~\ref{011}, where we have \(r_1=hypernym\) and \(r_2=r_3=derivationally\_related\_form\). In this case, we see that the composition pattern given by \(\mathcal{O}(\textbf{r2})\mathcal{O}(\textbf{r1})\) is learned to have similar scaling factor \(|Q|\) and rotation magnitude \(\psi\) with \(r_2\), and the rotation axes \((\theta, \phi)\) are also partially aligned.

\begin{figure}[htbp]
\centering
\begin{minipage}{3.5cm}
\centering
\includegraphics[width=3.5cm]{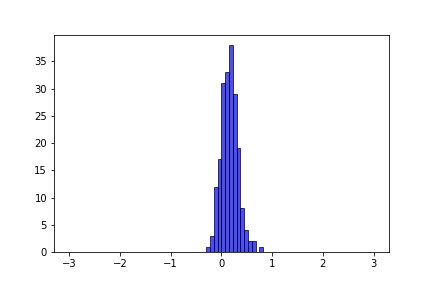}
\subcaption{\tiny{\(Q(\mathcal{O}(\textbf{r2})\mathcal{O}(\textbf{r1}))-Q(\mathcal{O}(\textbf{r2}))\)}}
\end{minipage}
\begin{minipage}{3.5cm}
\centering
\includegraphics[width=3.5cm]{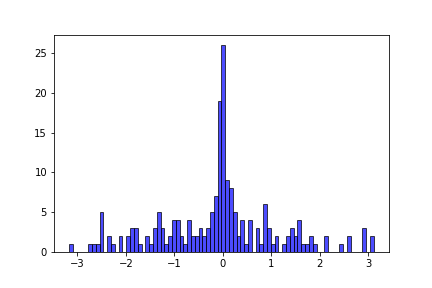}
\subcaption{\tiny{\(\psi(\mathcal{O}(\textbf{r2})\mathcal{O}(\textbf{r1}))-\psi(\mathcal{O}(\textbf{r2}))\)}}
\end{minipage}
\\
\begin{minipage}{3.5cm}
\centering
\includegraphics[width=3.5cm]{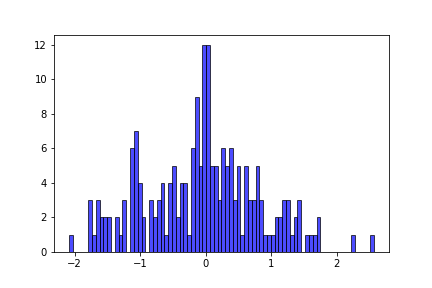}
\subcaption{\tiny{\(\theta(\mathcal{O}(\textbf{r2})\mathcal{O}(\textbf{r1}))-\theta(\mathcal{O}(\textbf{r2}))\)}}
\end{minipage}
\begin{minipage}{3.5cm}
\centering
\includegraphics[width=3.5cm]{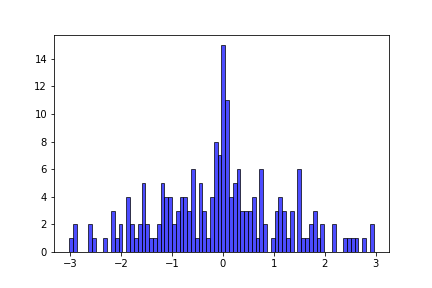}
\subcaption{\tiny{\(\phi(\mathcal{O}(\textbf{r2})\mathcal{O}(\textbf{r1}))-\phi(\mathcal{O}(\textbf{r2}))\)}}
\end{minipage}
\caption{Geometric interpretation of composition patterns in \textbf{Situation 4}.}
\label{011}
\end{figure}

\textbf{Situation 5:} \(r_1\), \(r_2\) and \(r_3\) are mutually different relations, e.g., the triangle in red color with three triplets set: (\(Trade(VB)\), \(derivationally\_related\_form\), \(Selling\)), (\(Selling\), \(hypernym\), \(mercantilism\)) and (\(Trade(VB)\), \(synset\_domain\_topic\_of\), \(mercantilism\)). Here we compare the difference between the embedding of a composite relation and the embedding calculated by multiplying each relation in the relation path. Related distributions are shown in Supplementary Figure~\ref{105}, where we have \(r_1=derivationally\_related\_form\), \(r_2=hypernym\) and \(r_3=synset\_domain\_topic\_of\). The reason for relatively large dispersion here is discussed in the main text, i.e., the majority of triplets exemplify \(derivationally\_related\_form \text{ }\Lambda \text{ } hypernym \Rightarrow derivationally\_related\_form\) (See also the discussion above of \textbf{Situation 3}), while only a small portion have the pattern \(derivationally\_related\_form \text{ } \Lambda \text{ } hypernym \Rightarrow synset\_domain\_topic\_of\). 

\begin{figure}[htbp]
\centering
\begin{minipage}{3.5cm}
\centering
\includegraphics[width=3.5cm]{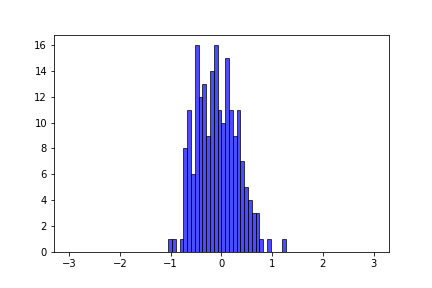}
\subcaption{\tiny{\(Q(\mathcal{O}(\textbf{r2})\mathcal{O}(\textbf{r1}))-Q(\mathcal{O}(\textbf{r3}))\)}}
\end{minipage}
\begin{minipage}{3.5cm}
\centering
\includegraphics[width=3.5cm]{images/wn18rr_derivationally_related_form_hypernym_synset_domain_topic_of_composition_theta_r.png}
\subcaption{\tiny{\(\psi(\mathcal{O}(\textbf{r2})\mathcal{O}(\textbf{r1}))-\psi(\mathcal{O}(\textbf{r3}))\)}}
\end{minipage}
\\
\begin{minipage}{3.5cm}
\centering
\includegraphics[width=3.5cm]{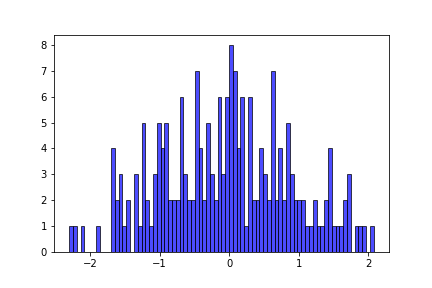}
\subcaption{\tiny{\(\theta(\mathcal{O}(\textbf{r2})\mathcal{O}(\textbf{r1}))-\theta(\mathcal{O}(\textbf{r3}))\)}}
\end{minipage}
\begin{minipage}{3.5cm}
\centering
\includegraphics[width=3.5cm]{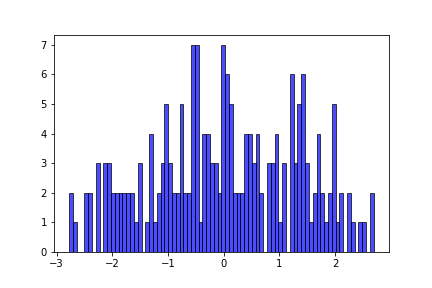}
\subcaption{\tiny{\(\phi(\mathcal{O}(\textbf{r2})\mathcal{O}(\textbf{r1}))-\phi(\mathcal{O}(\textbf{r3}))\)}}
\end{minipage}
\caption{Geometric interpretation of composition patterns in \textbf{Situation 5}.}
\label{105}
\end{figure}

\section{Link prediction results on WN18 dataset}

In the main text, we report the MRR and H@10 performance on WN18RR, FB15K237 and YAGO3-10. Here, we also report our model's performance on WN18 (Supplementary Table~\ref{WN18}), from which \textbf{inverse} relations are extracted by the demonstration of geometric interpretation of DensE (main text Figure~3).

\begin{table}[]
\centering
\caption{Performance comparison on WN18. The performances of RotatE are obtained from the original paper.}
 \begin{tabular}{ l c c c c c }
  \toprule
  
Model & MR & MRR	&H@1	&H@3&	H@10\\
  \midrule
RotatE	&	309	&	0.949	&	0.944	&	0.952	&	0.959		\\
DensE	&	285	&	0.950	&	0.945	&	0.954	&	0.959	\\
  \bottomrule
  \bottomrule
  \end{tabular} 
\label{WN18}

\end{table}

\section{Variance of the prediction performance}

The mean values and corresponding variance of MRR on WN18, WN18RR, FB15k-237 and YAGO3-10 datasets are shown in Supplementary Table~\ref{varience}. The results are obtained by training DensE with five different random seeds, showing that the prediction performance of DensE is relatively stable. 

\begin{table*}[!h]
\centering
\caption{The mean values and variance of MRR on WN18, WN18RR, FB15k-237 and YAGO3-10 datasets.}

 \begin{tabular}{ l c c c c  }
  \toprule
   ~ & WN18 & WN18RR & FB15k-237  & YAGO3-10 \\  
  \midrule
  MRR  & 0.950\(\pm\)0.001    & 0.492\(\pm\)0.001  & 0.351\(\pm\)0.001   & 0.541\(\pm\)0.001  \\ 
 \bottomrule
\end{tabular} 
\label{varience}
\end{table*}



\bibliography{emnlp2021}

\begin{thebibliography}{30}
\expandafter\ifx\csname natexlab\endcsname\relax\def\natexlab#1{#1}\fi

\bibitem[{Bala{\v{z}}evi{\'c} et~al.(2019)Bala{\v{z}}evi{\'c}, Allen, and
  Hospedales}]{balavzevic2019tucker}
Ivana Bala{\v{z}}evi{\'c}, Carl Allen, and Timothy~M Hospedales. 2019.
\newblock Tucker: Tensor factorization for knowledge graph completion.
\newblock \emph{arXiv preprint arXiv:1901.09590}.

\bibitem[{Bordes et~al.(2013)Bordes, Usunier, Garcia-Duran, Weston, and
  Yakhnenko}]{bordes2013translating}
Antoine Bordes, Nicolas Usunier, Alberto Garcia-Duran, Jason Weston, and Oksana
  Yakhnenko. 2013.
\newblock Translating embeddings for modeling multi-relational data.
\newblock In \emph{Advances in neural information processing systems}, pages
  2787--2795.

\bibitem[{Dettmers et~al.(2018)Dettmers, Minervini, Stenetorp, and
  Riedel}]{dettmers2018convolutional}
Tim Dettmers, Pasquale Minervini, Pontus Stenetorp, and Sebastian Riedel. 2018.
\newblock Convolutional 2d knowledge graph embeddings.
\newblock In \emph{Thirty-Second AAAI Conference on Artificial Intelligence}.

\bibitem[{Gao et~al.(2020)Gao, Sun, Shan, Lin, and Wang}]{gao2020rotate3d}
Chang Gao, Chengjie Sun, Lili Shan, Lei Lin, and Mingjiang Wang. 2020.
\newblock Rotate3d: Representing relations as rotations in three-dimensional
  space for knowledge graph embedding.
\newblock In \emph{Proceedings of the 29th ACM International Conference on
  Information \& Knowledge Management}, pages 385--394.

\bibitem[{Ji et~al.(2020)Ji, Pan, Cambria, Marttinen, and Yu}]{ji2020survey}
Shaoxiong Ji, Shirui Pan, Erik Cambria, Pekka Marttinen, and Philip~S Yu. 2020.
\newblock A survey on knowledge graphs: Representation, acquisition and
  applications.
\newblock \emph{arXiv preprint arXiv:2002.00388}.

\bibitem[{Jia(2019)}]{jia2019quaternions}
Yan-Bin Jia. 2019.
\newblock Quaternions.
\newblock \emph{Com S}, 477:577.

\bibitem[{Lacroix et~al.(2018)Lacroix, Usunier, and
  Obozinski}]{Lacroix2018CanonicalTD}
Timoth{\'e}e Lacroix, Nicolas Usunier, and Guillaume Obozinski. 2018.
\newblock Canonical tensor decomposition for knowledge base completion.
\newblock In \emph{ICML}.

\bibitem[{Lin et~al.(2019)Lin, Chen, Chen, and Ren}]{Lin2019KagNetKG}
Bill~Yuchen Lin, Xinyue Chen, Jamin Chen, and Xiang Ren. 2019.
\newblock Kagnet: Knowledge-aware graph networks for commonsense reasoning.
\newblock In \emph{EMNLP/IJCNLP}.

\bibitem[{Mahdisoltani et~al.(2013)Mahdisoltani, Biega, and
  Suchanek}]{mahdisoltani2013yago3}
Farzaneh Mahdisoltani, Joanna Biega, and Fabian~M Suchanek. 2013.
\newblock Yago3: A knowledge base from multilingual wikipedias.

\bibitem[{Miller(1995)}]{miller1995wordnet}
George~A Miller. 1995.
\newblock Wordnet: a lexical database for english.
\newblock \emph{Communications of the ACM}, 38(11):39--41.

\bibitem[{Nathani et~al.(2019)Nathani, Chauhan, Sharma, and
  Kaul}]{nathani2019learning}
Deepak Nathani, Jatin Chauhan, Charu Sharma, and Manohar Kaul. 2019.
\newblock Learning attention-based embeddings for relation prediction in
  knowledge graphs.
\newblock In \emph{Proceedings of the 57th Annual Meeting of the Association
  for Computational Linguistics}, pages 4710--4723.

\bibitem[{Nguyen et~al.(2017)Nguyen, Nguyen, Nguyen, and
  Phung}]{nguyen2017novel}
Dai~Quoc Nguyen, Tu~Dinh Nguyen, Dat~Quoc Nguyen, and Dinh Phung. 2017.
\newblock A novel embedding model for knowledge base completion based on
  convolutional neural network.
\newblock \emph{arXiv preprint arXiv:1712.02121}.

\bibitem[{Nickel et~al.(2016)Nickel, Rosasco, and
  Poggio}]{nickel2016holographic}
Maximilian Nickel, Lorenzo Rosasco, and Tomaso Poggio. 2016.
\newblock Holographic embeddings of knowledge graphs.
\newblock In \emph{Thirtieth Aaai conference on artificial intelligence}.

\bibitem[{Nickel et~al.(2011)Nickel, Tresp, and Kriegel}]{nickel2011three}
Maximilian Nickel, Volker Tresp, and Hans-Peter Kriegel. 2011.
\newblock A three-way model for collective learning on multi-relational data.
\newblock In \emph{Icml}, volume~11, pages 809--816.

\bibitem[{Ruffinelli et~al.(2019)Ruffinelli, Broscheit, and
  Gemulla}]{ruffinelli2019you}
Daniel Ruffinelli, Samuel Broscheit, and Rainer Gemulla. 2019.
\newblock You can teach an old dog new tricks! on training knowledge graph
  embeddings.
\newblock In \emph{International Conference on Learning Representations}.

\bibitem[{Schlichtkrull et~al.(2018)Schlichtkrull, Kipf, Bloem, Van Den~Berg,
  Titov, and Welling}]{schlichtkrull2018modeling}
Michael Schlichtkrull, Thomas~N Kipf, Peter Bloem, Rianne Van Den~Berg, Ivan
  Titov, and Max Welling. 2018.
\newblock Modeling relational data with graph convolutional networks.
\newblock In \emph{European Semantic Web Conference}, pages 593--607. Springer.

\bibitem[{Sun et~al.(2019)Sun, Deng, Nie, and Tang}]{sun2019rotate}
Zhiqing Sun, Zhi-Hong Deng, Jian-Yun Nie, and Jian Tang. 2019.
\newblock Rotate: Knowledge graph embedding by relational rotation in complex
  space.
\newblock In \emph{The Seventh International Conference on Learning
  Representations}.

\bibitem[{Toutanova and Chen(2015{\natexlab{a}})}]{Toutanova2015ObservedVL}
Kristina Toutanova and Danqi Chen. 2015{\natexlab{a}}.
\newblock Observed versus latent features for knowledge base and text
  inference.

\bibitem[{Toutanova and Chen(2015{\natexlab{b}})}]{toutanova2015observed}
Kristina Toutanova and Danqi Chen. 2015{\natexlab{b}}.
\newblock Observed versus latent features for knowledge base and text
  inference.
\newblock In \emph{Proceedings of the 3rd Workshop on Continuous Vector Space
  Models and their Compositionality}, pages 57--66.

\bibitem[{Trouillon et~al.(2016{\natexlab{a}})Trouillon, Welbl, Riedel,
  Gaussier, and Bouchard}]{trouillon2016complex}
Th{\'e}o Trouillon, Johannes Welbl, Sebastian Riedel, {\'E}ric Gaussier, and
  Guillaume Bouchard. 2016{\natexlab{a}}.
\newblock Complex embeddings for simple link prediction.
\newblock In \emph{International Conference on Machine Learning}, pages
  2071--2080.

\bibitem[{Trouillon et~al.(2016{\natexlab{b}})Trouillon, Welbl, Riedel,
  Gaussier, and Bouchard}]{Trouillon2016ComplexEF}
Th{\'e}o Trouillon, Johannes Welbl, Sebastian Riedel, {\'E}ric Gaussier, and
  Guillaume Bouchard. 2016{\natexlab{b}}.
\newblock Complex embeddings for simple link prediction.
\newblock \emph{ArXiv}, abs/1606.06357.

\bibitem[{Wang et~al.(2017)Wang, Gemulla, and
  Li}]{DBLP:journals/corr/abs-1709-04808}
Yanjie Wang, Rainer Gemulla, and Hui Li. 2017.
\newblock \href {http://arxiv.org/abs/1709.04808} {On multi-relational link
  prediction with bilinear models}.
\newblock \emph{CoRR}, abs/1709.04808.

\bibitem[{Wang et~al.(2018)Wang, Ruffinelli, Gemulla, Broscheit, and
  Meilicke}]{wang2018evaluating}
Yanjie Wang, Daniel Ruffinelli, Rainer Gemulla, Samuel Broscheit, and Christian
  Meilicke. 2018.
\newblock On evaluating embedding models for knowledge base completion.
\newblock \emph{arXiv preprint arXiv:1810.07180}.

\bibitem[{Xu and Li(2019)}]{xu2019relation}
Canran Xu and Ruijiang Li. 2019.
\newblock Relation embedding with dihedral group in knowledge graph.
\newblock \emph{arXiv preprint arXiv:1906.00687}.

\bibitem[{Yang et~al.(2019)Yang, Wang, Liu, Liu, Lyu, Wu, She, and
  Li}]{yang-etal-2019-enhancing-pre}
An~Yang, Quan Wang, Jing Liu, Kai Liu, Yajuan Lyu, Hua Wu, Qiaoqiao She, and
  Sujian Li. 2019.
\newblock \href {https://doi.org/10.18653/v1/P19-1226} {Enhancing pre-trained
  language representations with rich knowledge for machine reading
  comprehension}.
\newblock In \emph{Proceedings of the 57th Annual Meeting of the Association
  for Computational Linguistics}, pages 2346--2357, Florence, Italy.
  Association for Computational Linguistics.

\bibitem[{Yang et~al.(2014)Yang, Yih, He, Gao, and Deng}]{yang2014embedding}
Bishan Yang, Wen-tau Yih, Xiaodong He, Jianfeng Gao, and Li~Deng. 2014.
\newblock Embedding entities and relations for learning and inference in
  knowledge bases.
\newblock \emph{arXiv preprint arXiv:1412.6575}.

\bibitem[{Yang et~al.(2020)Yang, Sha, and Hong}]{Yang2020AGF}
Tong Yang, Long Sha, and Pengyu Hong. 2020.
\newblock \href {https://doi.org/10.1145/3340531.3411875} {Nage: Non-abelian
  group embedding for knowledge graphs}.
\newblock In \emph{Proceedings of the 29th ACM International Conference on
  Information \& Knowledge Management}, CIKM '20, pages 1735--1742, New York,
  NY, USA. Association for Computing Machinery.

\bibitem[{Zhang et~al.(2019{\natexlab{a}})Zhang, Tay, Yao, and
  Liu}]{zhang2019quaternion}
Shuai Zhang, Yi~Tay, Lina Yao, and Qi~Liu. 2019{\natexlab{a}}.
\newblock Quaternion knowledge graph embeddings.
\newblock In \emph{Advances in Neural Information Processing Systems}, pages
  2731--2741.

\bibitem[{Zhang et~al.(2020)Zhang, Cai, Zhang, and Wang}]{zhang2020learning}
Zhanqiu Zhang, Jianyu Cai, Yongdong Zhang, and Jie Wang. 2020.
\newblock Learning hierarchy-aware knowledge graph embeddings for link
  prediction.
\newblock In \emph{AAAI}, pages 3065--3072.

\bibitem[{Zhang et~al.(2019{\natexlab{b}})Zhang, Han, Liu, Jiang, Sun, and
  Liu}]{Zhang2019ERNIEEL}
Zhengyan Zhang, Xu~Han, Zhiyuan Liu, Xin Jiang, Maosong Sun, and Qun Liu.
  2019{\natexlab{b}}.
\newblock Ernie: Enhanced language representation with informative entities.
\newblock \emph{ArXiv}, abs/1905.07129.

\end{thebibliography}
\bibliographystyle{acl_natbib}

\end{document}